\documentclass{article} 
\usepackage{iclr2023_workshop,times}

\usepackage{amsmath,amsfonts,bm}

\def\eqref#1{equation~\ref{#1}}

\def\1{\bm{1}}

\DeclareMathAlphabet{\mathsfit}{\encodingdefault}{\sfdefault}{m}{sl}
\SetMathAlphabet{\mathsfit}{bold}{\encodingdefault}{\sfdefault}{bx}{n}

\usepackage{hyperref}
\usepackage{url}

\usepackage[utf8]{inputenc} 
\usepackage[T1]{fontenc}    
\usepackage{booktabs}       
\usepackage{amsfonts}       
\usepackage{nicefrac}       
\usepackage{microtype}      
\usepackage{xcolor}         
\usepackage{caption}
\usepackage{subcaption}
\usepackage{graphicx}
\usepackage{siunitx}
\usepackage{xspace}

\newcommand{\patches}{patches}
\newcommand{\patch}{patch}

\newcommand{\ghostZone}{ghost zone}
\newcommand\thickness[1]{$#1$}
\newcommand\CTFD{\textsc{C3FD}\xspace}

\title{Scientific Computing Algorithms to Learn Enhanced Scalable Surrogates for Mesh Physics} 

\author{Brian R. Bartoldson$^\dagger$, 
Yeping Hu$^\dagger$\thanks{Equal contribution. 
Corresponding authors: \texttt{bartoldson@llnl.gov}, \texttt{nguyen97@llnl.gov}.
}, 
Amar Saini$^\dagger$\rlap{,}$^\ast$ Jose Cadena$^\dagger$\rlap{,}$^\ast$
Yucheng Fu$^\ddagger$, Jie Bao$^\ddagger$, \\
\textbf{Zhijie Xu$^\ddagger$, Brenda Ng$^\dagger$, \& Phan Nguyen$^\dagger$}
\\
$^\dagger$Lawrence Livermore National Laboratory ~~ $^\ddagger$Pacific Northwest National Laboratory \\
}

\iclrfinalcopy 
\begin{document}

\maketitle

\begin{abstract}

Data-driven modeling approaches can produce fast surrogates to study large-scale physics problems. Among them, graph neural networks (GNNs) that operate on mesh-based data are desirable because they possess inductive biases that promote physical faithfulness, but hardware limitations have precluded their application to large computational domains. We show that it is \textit{possible} to train a class of GNN surrogates on 3D meshes. We scale MeshGraphNets (MGN), a subclass of GNNs for mesh-based physics modeling, via our domain decomposition approach to facilitate training that is mathematically equivalent to training on the whole domain under certain conditions. With this, we were able to train MGN on meshes with \textit{millions} of nodes to generate computational fluid dynamics (CFD) simulations. Furthermore, we show how to enhance MGN via higher-order numerical integration, which can reduce MGN's error and training time. We validated our methods on an accompanying dataset of 3D $\text{CO}_2$-capture CFD simulations on a 3.1M-node mesh. This work presents a practical path to scaling MGN for real-world applications. 

\end{abstract}

\section{Introduction}

Understanding physical systems and engineering processes often requires extensive numerical simulations of their underlying models. However, these simulations are typically computationally expensive to generate, which can hinder their applicability to large-scale problems. In such cases, surrogates can be used to emulate the original models and produce computationally efficient simulations. 

Surrogate approaches can suggest efficiency-fidelity tradeoffs~\citep{Willard2020Survey,Wang2021PhysicsGuidedDL}. For example, while physically faithful and generalizable to unseen domains, mesh-based GNNs offer smaller speedups (11--290x) \citep{pfaff2020learning} than those of less generalizable CNNs (125--716x) \citep{kim2019}. Moreover, hardware limitations and memory requirements of mesh-based GNNs prevent their application to large graphs. To the best of our knowledge, they have not been applied to graphs with more than 20K nodes \citep{fortunato2022multiscale}. Despite this tradeoff, many applications of surrogates require \textit{both} generalization to unseen domains and efficient processing on large datasets. 

Thus, we seek to improve the applicability of GNNs without harming the inductive biases critical to their accuracy by building on scientific computing algorithms. Focusing on MeshGraphNets (MGN) \citep{pfaff2020learning}, we enhance MGN learning via domain decomposition and higher-order numerical integration---methods that fall into a framework we call \textbf{Scientific Computing Algorithms to Learn Enhanced Scalable Surrogates (SCALES2)}. Our contributions are the following:

\begin{enumerate}
    
    \item We train GNN surrogates on CFD meshes with millions of nodes, roughly 100 times larger than the largest meshes used in prior work \citep{pfaff2020learning,fortunato2022multiscale}.
    
    \item We present patch training, a domain decomposition method for training on large meshes, and conditions for its mathematical equivalence to training on a large mesh directly.
    
    \item We show that MGN accuracy and efficiency can be greatly improved by using higher-order numerical integration schemes that can augment surrogate faithfulness.

    \item We release \CTFD, a large-scale dataset of Carbon Capture CFD simulations on 3D meshes containing 3.1M nodes, to facilitate benchmarking of graph-based surrogates.

\end{enumerate}

\section{Approach}

\subsection{MeshGraphNets}

MGN uses message-passing (MP) GNNs to learn from mesh-based simulations. In each MP step, MGN updates each edge's features by aggregating the features of its incident nodes, and then updates each node's features by aggregating the features of its incident edges. After $k$ iterations, each node and edge will be influenced by its $k$-hop neighborhood. The final features of each node are then used to predict the node's next state using the following update rule: $y_{t+h} = y_t + \text{MGN}(y_t)$, where $y_t$ is the system's numerically estimated state at time $t$ and $h$ is the timestep size.

\begin{figure*}[t]
	\centering
	\includegraphics[scale=0.055]{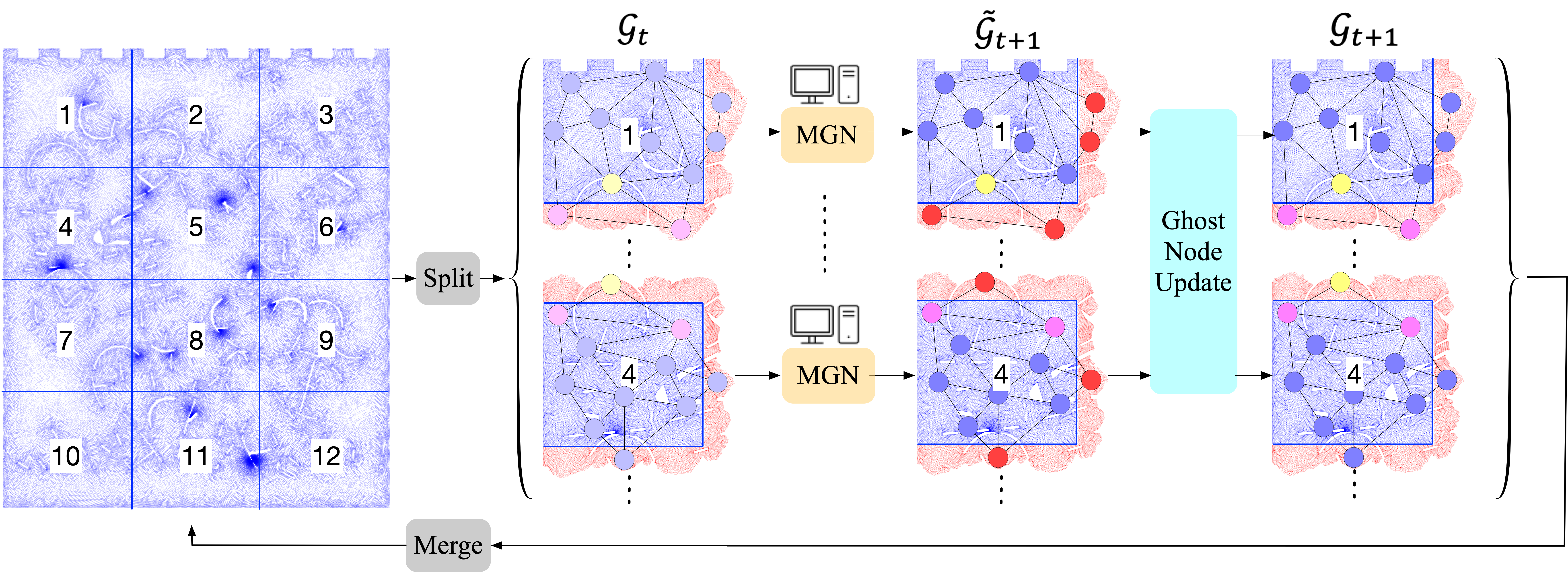}
	\caption{\textbf{Scaling MGN via domain decomposition.} We partition graph $\mathcal{G}_t$ into $P$ subdomains (squares), augment the subdomains with ghost nodes to get overlapping patches (e.g., Patch1 and Patch4 have ghost nodes from each other, shown as pink and yellow nodes), then run MGN on the patches $\mathcal{G}^{(p)}_t$ to obtain $\mathcal{\tilde{G}}^{(p)}_{t+1}$.
 For inference, nodes with incorrect states (red)---due to ``incomplete'' message-passing with a subset of their neighbors---are updated by communicating across patches to obtain $\mathcal{G}^{(p)}_{t+1}$. When training, ghost node updates are not needed as we predict only the next timestep.}
	\label{fig:framework}
\end{figure*}

\subsection{Domain decomposition for scaling}
\label{sec:DD}

We develop a domain decomposition framework that allows mesh-based GNN surrogates to be trained on large domains (see Figure~\ref{fig:framework}). Domain decomposition is useful when application of a surrogate to a domain produces out-of-memory errors on the hardware used for inference or learning. Broadly, it divides a simulation problem defined on a domain into a jointly equivalent (or similar) set of problems, each of which is defined on its own subdomain \citep{tang2021review}. 

For our datasets, we create evenly-sized subdomains by dividing each dimension of the domain into equal parts. Inspired by overlapping domain decomposition approaches \citep{tang2021review}, we augment each subdomain with its $k$-hop neighborhood, which we call the \ghostZone{} with width \thickness{k}. We call the union of the subdomain and its \ghostZone{} a \patch{}. Because each subdomain node can access its $k$-hop neighborhood in its \patch{}, an MGN with $m$ MP steps, where $m \leq $ \thickness{k}, will update each subdomain node as if the entire domain had been available. Thus, training on \patches{} is equivalent to training on the full domain, under three conditions: (1) $m \leq $ \thickness{k}; (2) we only compute the loss on the subdomain nodes (excluding the ghost nodes); and (3) we accumulate the gradients over multiple batches until all patches in the domain have been visited before updating the weights. In practice, we violate the third condition by updating the weights after a random subset of a domain's patches is seen---we call the resulting procedure ``patch training''. 

\subsection{Higher-order integration for enhanced learning}
\label{sec:int_approach}

MGN's update rule is based on the forward Euler (FE) method, which introduces first-order global error and second-order local truncation error by ignoring quadratic and higher-order terms in the following Taylor series: $y(t+h) = y(t) + h f(y(t)) + \mathcal{O}(h^2)$, where $y(t)$ is the true state of a system at time $t$. Accordingly, MGN's prediction error is $\left|y(t+h)-y_{t+h}\right| = \left| [h f(y(t))-\mathrm{MGN}(y(t))] + \mathcal{O}(h^2) \right|$, which can be decomposed into a prediction error of the (scaled) time derivative, $[hf(y) - \text{MGN}(y)]$, and a second-order local error, $\mathcal{O}(h^2)$.\footnote{In practice, this analysis is not exact because MGN error uses CFD data that only approximates $y(t+h)$.} 

While MGN may learn to offset the $\mathcal{O}(h^2)$ term when learning to estimate the time derivative, local error can be reduced by changing how MGN updates $y_t$ to $y_{t+h}$, increasing the benefit of correctly predicting the time derivative. Specifically, we replace FE with Heun's second- and third-order methods~\citep{heun1900neue,Butcher1996numerical}, which have smaller local errors ($\mathcal{O}(h^3)$ and $\mathcal{O}(h^4)$, respectively). In Appendix~\ref{sec:taylor}, we hypothesize that this can speed up and enhance learning. 

\section{Datasets}

\textbf{CylinderFlow} CylinderFlow contains 2D simulations of inviscid, incompressible flow around a cylinder, generated by numerical solvers with Navier-Stokes equations on an Eulerian mesh~\citep{pfaff2020learning}. We train surrogates to predict the fluid momenta at each node of the mesh. 

\textbf{\CTFD} Our Carbon Capture CFD (\CTFD) dataset contains STAR-CCM+ simulations of a $\text{CO}_2$-capturing solvent flowing across packing structures in a column. The dataset contains 50 2D and 50 3D simulations, each spanning 500 timesteps. The 2D and 3D simulation meshes have 150K and 3.1M nodes, respectively. Each node contains information about its physical location, type, liquid volume fraction (a quantity related to capture efficiency), momentum per unit volume, and contact angle. We train surrogates to predict liquid volume fraction and momentum at each node. Additional dataset details and visualizations are available in Appendix~\ref{sec:data_details}. 
The \CTFD dataset is available at \url{https://doi.org/10.25584/1963908}.

\section{Experiments}

As a sanity check, we first replicate experiments from \citet{pfaff2020learning} on the full 2D \CTFD domain. In Figures~\ref{fig:heun} and \ref{fig:rollout}, we show that we can train MGNs to low error and generate 500-timestep rollouts that are similar to the ground truth. Further, we show that MGNs trained on the 2D \CTFD domain generalize to a distinct 2D domain (i.e., a new slice of the 3D domain).

These initial experiments on 2D \CTFD required roughly 50 GB of accelerator memory to apply the original MGN. Applying MGN to larger datasets like the 3D \CTFD dataset, which requires roughly 1 TB of accelerator memory, is not possible with commonly available hardware. Thus, aiming to make mesh-based GNN surrogates more efficient, we test algorithms to scale and enhance their learning. Ultimately, our patch training successfully scales MGN training to learn from the 3D \CTFD data. We also show that MGN training can be enhanced by using higher-order integration.

\subsection{Patch training is key to scaling surrogates}
\label{sec:DD_exp}

In Section~\ref{sec:DD}, we outlined patch training. However, it remains unclear how mesh surrogates like MGN perform on very large domains. Training on a full domain with domain decomposition may be wasteful when the domain is very large, because each patch in the domain must be visited before the weights are updated. We explore if it may be just as effective to use patch training, which updates the weights after visiting a small sample of a domain's patches. 

\begin{figure}[t]
\centering
\begin{subfigure}{.47\textwidth}
  \centering
  \includegraphics[width=\textwidth]{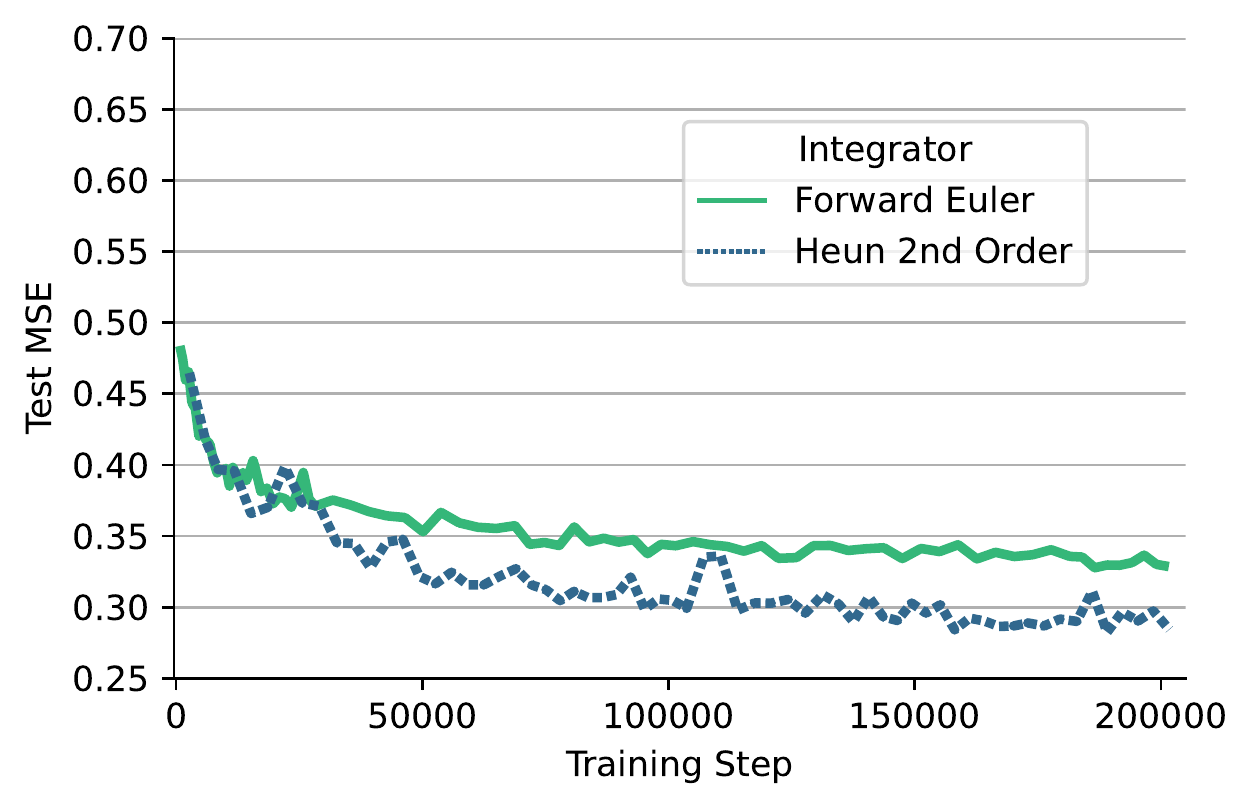}
\end{subfigure}
\hfill
\begin{subfigure}{.493\textwidth}
  \centering
  \includegraphics[width=\textwidth]{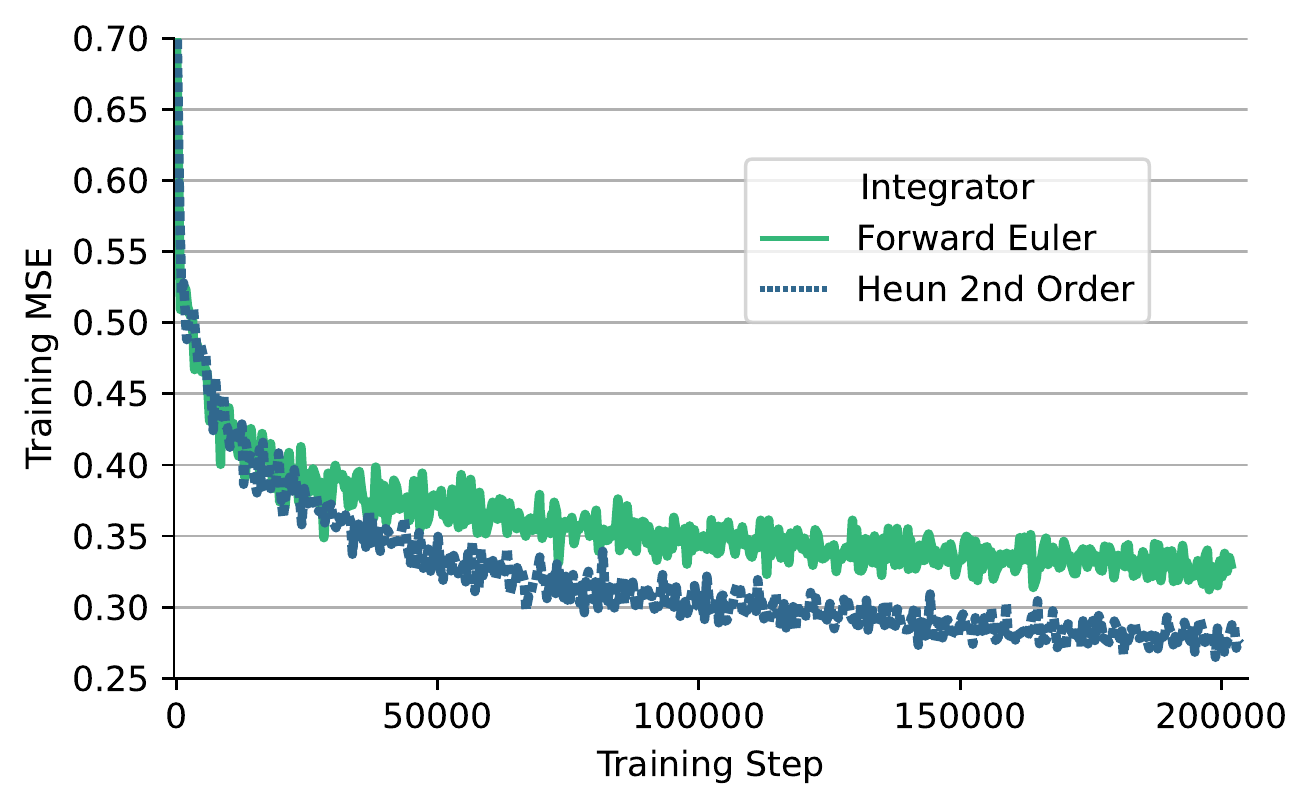}
\end{subfigure}
\caption{\textbf{Patch training on 3D \CTFD}. Experiments run on data with 3.1M nodes per timestep.}
\label{fig:3d}
\end{figure}

Unlike training on smaller domains and then generalizing to a large test domain (Figure~\ref{fig:subset}), we find that patch training induces no performance drop relative to training on the large domain directly. This holds for CylinderFlow (Figure~\ref{fig:cf}) and 2D \CTFD (Figure~\ref{fig:patch}). Based on this, we use patch training to efficiently train MGNs on a 3D mesh with millions of nodes (3D \CTFD) for the first time (Figure~\ref{fig:3d}). To optimize this model for practical deployment, further work (e.g., hyperparameter tuning) remains. We provide additional experiment details and results in Appendices~\ref{sec:experiment_details} and \ref{sec:supporting}, respectively.

\subsection{Higher-order integration enhances learning}
\label{sec:int_exp}

Following discussion from Section~\ref{sec:int_approach}, we want to investigate: How significant are the $\mathcal{O}(h^2)$ truncation errors from FE? If significant, how effective is MGN in learning to correct these errors? While it is clear that higher-order integration methods should reduce the truncation errors, these higher-order integrators may not provide benefits if MGN can easily correct for these errors.

\begin{figure}[t]
\centering
\begin{subfigure}{.47\textwidth}
  \centering
  \includegraphics[width=\textwidth]{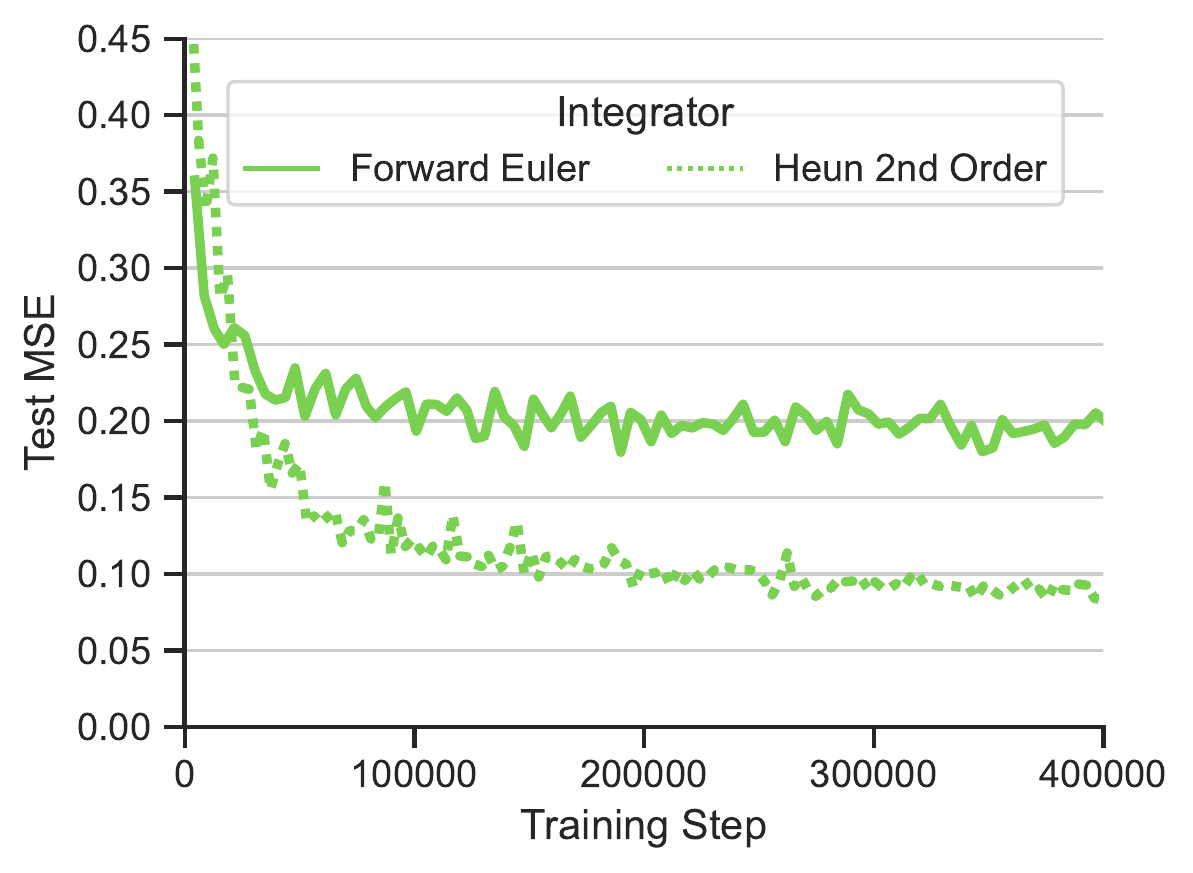}
\end{subfigure}
\hfill
\begin{subfigure}{.482\textwidth}
  \centering
  \includegraphics[width=\textwidth]{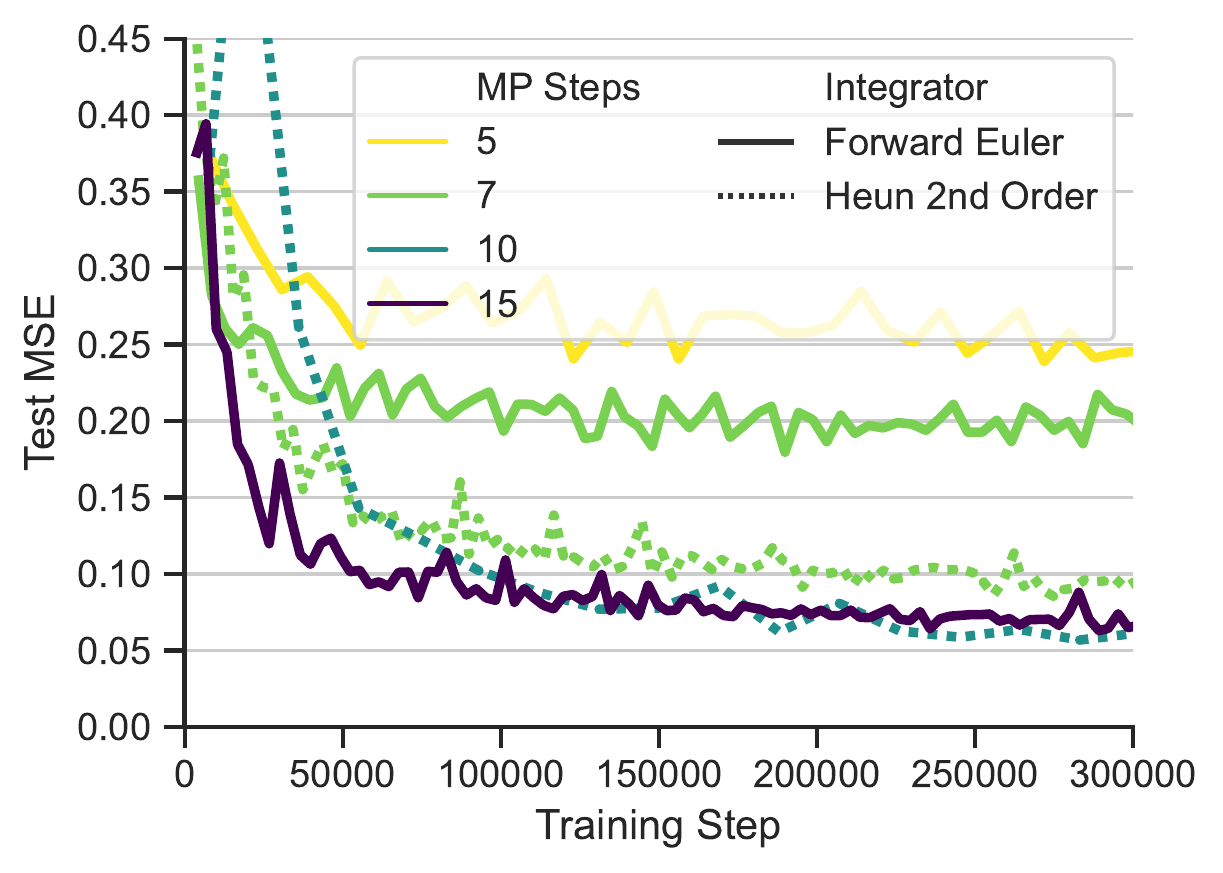}
\end{subfigure}
\caption{\textbf{Higher-order integration enhances learning}. Experiments run on 2D \CTFD data with 150K nodes per timestep. (\textit{Left}) An MGN using Heun's second-order method (H2) with 7 message-passing (MP) steps is able to halve the error---relative to forward Euler (FE)---on both test (as shown) and training (see Figure~\ref{fig:heun_train}) data. (\textit{Right}) H2 with 7 MP steps is comparable to FE with 15 MP steps, despite the benefits associated with more MP steps. H2 with 10 MP steps can outperform FE with 15 MP steps. In Figure~\ref{fig:int_stable}, we show higher-order integration's benefits are stable across training runs.}
\label{fig:heun}
\end{figure}

Using the 2D \CTFD data, we train MGNs using three different integrators: FE, Heun's second-order method (H2), and Heun's third-order method (H3). Comparing FE and H2, we see that test error can be roughly halved with H2 (Figure~\ref{fig:heun}~\textit{left}). Using more MP steps generally improves surrogate performance, so we also compare the integrators across a range of MP steps (Figure~\ref{fig:heun}~\textit{right}). The H2-surrogate using 7 MP steps achieves roughly the same error as the FE-surrogate using 15 steps. Furthermore, H2 with 10 MP steps reaches even lower error than FE with 15 MP steps. This suggests that higher-order integration is an important tool for improving MGN performance, with the added bonus of requiring fewer MP steps to achieve the same level of accuracy. We provide additional discussion and results in Appendix~\ref{sec:supporting} and review related work in Appendix~\ref{sec:related}.

\section{Discussion}

We introduced scientific computing algorithms (SCALES2) to enhance and scale training of GNN surrogates on mesh physics. Our approach enabled MGN to train on 3D meshes with millions of nodes. We also showed that higher-order integration can halve the error in our 2D experiments. To stimulate further research on scaling and improving surrogate simulation, we released \CTFD, a benchmark dataset with $\text{CO}_2$ capture CFD simulations on both 2D and 3D meshes (with 150K nodes and 3.1M nodes, respectively). As this study focuses on next-step predictions, it remains to be seen the overall improvement that SCALES2 may provide to longer rollouts on large domains; we discuss additional limitations and extensions to our work in Appendix~\ref{sec:limitations}.

\subsubsection*{Acknowledgments}
This work was performed under the auspices of the U.S. Department of Energy by Lawrence Livermore National Laboratory under Contract DE-AC52-07NA27344. This research used resources of the National Energy Research Scientific Computing Center, a DOE Office of Science User Facility supported by the Office of Science of the U.S. Department of Energy under Contract No. DE-AC02-05CH11231 using NERSC award BES-ERCAP0024437. This work was conducted as part of the Carbon Capture Simulation for Industry Impact (CCSI2) project.

LLNL Release Number (Paper): LLNL-CONF-845156. \\ PNNL Release Number (Dataset): PNNL-SA-182276.

\bibliography{references}
\bibliographystyle{iclr2023_workshop}

\appendix

\section{Higher-order integration analysis}
\label{sec:taylor} 

Integrators of sufficiently high order can practically eliminate the local truncation error, suggesting that MGN updates based on such integrators can produce extremely high accuracy solely through learning to compute the time derivative $f(y)$. Accordingly, we explore their use as a path toward improving surrogate performance. In Section~\ref{sec:int_exp} and Appendix~\ref{sec:heun_train_and_third}, we train surrogates that use forward Euler (FE) (a first-order method), and Heun's second- (H2) and third-order (H3) methods. 

The H2 method is as follows: $y_{t+h} = y_t + \frac{h}{2}(f(y_t) + f(y_t + hf(y_t)))$. Comparing this to FE (Equation~\ref{eq:FE_HUEN}), we find that H2 requires the MGN to also learn $f(y)$ as FE does, but H2 incurs lower truncation error. When only first-order derivative information is learnable in a physically consistent manner, higher-order integrators may make MGNs more physically faithful by reducing the scale of the truncation error relative to the scale of the error in the first-order derivative.

\begin{equation}
    \begin{split}
       {y(t+h)-y(t)} \; &=  \quad \underbrace{h f(y(t))}_\text{FE update} \quad \ \ \ +  \underbrace{\mathcal{O}(h^2)}_\text{truncation error} \\
                        &= \quad  \underbrace{\frac{h}{2}\Big(f\big(y(t)\big) + f\big(y(t)+hf(y(t))\big)\Big)}_\text{H2 update} \;\; +   \underbrace{\mathcal{O}(h^3)}_\text{truncation error} .
    \end{split}
    \label{eq:FE_HUEN}
\end{equation}

Using a state update based on H2, MGN's error can again (see Section~\ref{sec:int_approach}) be decomposed into two terms: an error based on MGN's ability to predict the (scaled) time derivative,
\begin{equation}
\Bigg[\frac{h}{2}\Big(f\big(y(t)\big) + f\big(y(t)+hf(y(t))\big)\Big) - \frac{1}{2}\Big(\text{MGN}(y(t)) + \text{MGN}\big(y(t)+\text{MGN}(y(t))\big)\Big)\Bigg],
    \label{eq:heun}
\end{equation}
and the $\mathcal{O}(h^3)$ term, a \textit{third-order} local truncation error. Note that while MGN still only needs to learn to predict the scaled time derivative, the benefit of $\text{MGN}(y(t))$ mimicking $hf(y)$ increases with an H2-based updated because the truncation error term is now smaller.

Therefore, higher-order integration may provide \textit{more accurate learning} (lower final error) by explicitly eliminating $\mathcal{O}(h^2)$ truncation errors that are not removable by FE-based training, wherein the original MGN is not guaranteed to learn how to offset the varying $\mathcal{O}(h^2)$ error. Whether these corrections can be learned depends on properties of the data and surrogate (e.g., if the second derivative is learnable, then such corrections may be achievable). We also note that higher-order integration may provide \textit{faster learning} by focusing learning on $f(y)$ (i.e., by providing a better signal-to-noise ratio). 

\section{Dataset details}
\label{sec:data_details}

\subsection{\CTFD}

\begin{figure}[t]
\centering
 \includegraphics[width=0.9\textwidth]{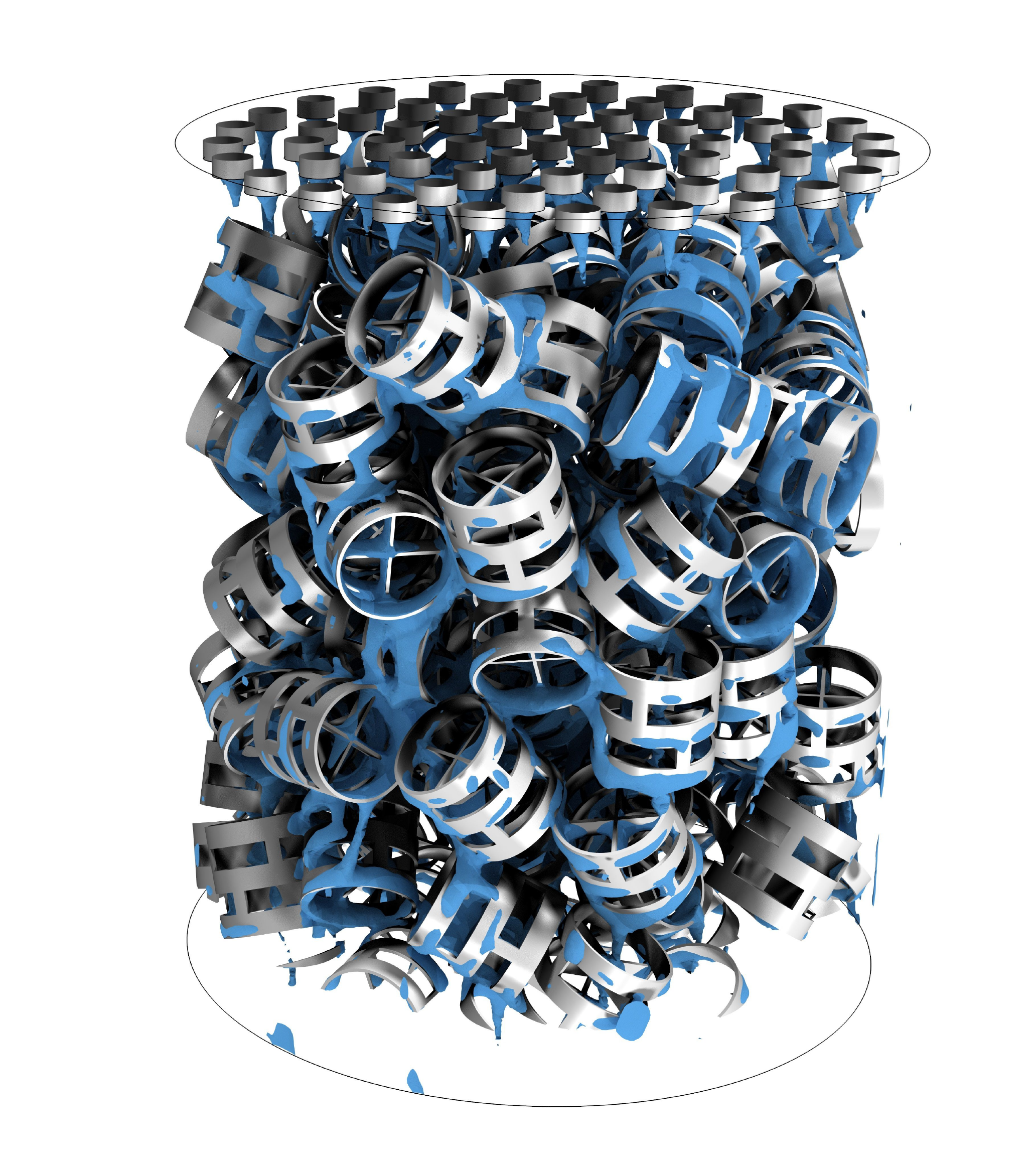}
\caption{Visualization of the 3D \CTFD domain, which is a 100mm wide and 200mm tall cylinder packed with pall rings. Gas enters from below and exits above, while liquid does the opposite.}
\label{fig:rcm}
\end{figure}

Figure~\ref{fig:rcm} shows the 3D \CTFD domain, which is a representative section of a bench-scale column of a $\text{CO}_2$-capture system. The 2D domain (see Figure~\ref{fig:framework} for an illustration) has 150K polygonal mesh grid points, while the 3D domain has 3.1M points (the exact number varies, as each simulation used one of three unique meshes). There are 50 3D simulations and 50 2D simulations. The different simulations correspond to different velocities for the liquid entering at the inlets at the top of the domain; other variables, such as the geometry of the domain, are held constant across the 50 simulations. Training and validation splits were created to allow validation to test extrapolation and interpolation. Validation simulations covered the four slowest, four fastest, and four intermediate inlet velocity simulations. 

\CTFD followed the setup described in \citet{fu2020investigation}: Eulerian simulations of the counter-current solvent and gas flow were generated by solving the continuity and momentum equations and using the volume-of-fluid method to model multiphase fluid separation. STAR-CCM+ was used to solve

\begin{equation}
\frac{\partial \rho}{\partial t}+\nabla \cdot \rho \mathbf{u}=0,
\end{equation}
\begin{equation}
\frac{\partial(\rho \mathbf{u})}{\partial t}+\nabla \cdot(\rho \mathbf{u u})=-\nabla p+\mu \nabla^{2} \mathbf{u}+\rho \mathbf{g}+\mathbf{F}_{\sigma},
\end{equation}
where $\mathbf{u}$ is velocity,  $\rho$ is density, $p$ is pressure, $\mathbf{g}$ is gravity, $\mu$  is viscosity,  and $\mathbf{F}_{\sigma}$ is the surface tension at the gas-liquid interface. Density and viscosity are calculated via the volume fraction average of the liquid ($\alpha$) and gas phase (1-$\alpha$). The following transport equation governs the evolution of $\alpha$:
\begin{equation}
\frac{\partial \alpha}{\partial t}+\nabla \cdot (\mathbf{u} \alpha)=0.
\end{equation}

The 2D simulation domain is a cross-section of the 3D domain, but it was solved on independently. For 2D, the solver used timestep size $h = 0.001 s$, and data was output at every tenth timestep. For 3D, the solver used $h = 0.0002 s$, and data was saved at every fiftieth timestep. The 2D and 3D data provided in \CTFD was therefore saved using $h = 0.01 s$. 

To create edges, we used SciPy's implementation of Delaunay triangulation~\citep{scipy}. Notably, we noticed some splinter-like cells after this procedure; further mesh refinement or filtering may be applied to resolve this, e.g., the procedure described by \citet{lienen2022learning}.

The following quantities are provided at each node of the computational domain: position, velocity, pressure, and liquid volume fraction, and boundary index. For velocity, a value is provided for each spatial dimension. Rather than include the momentum vector in the dataset explicitly, we compute it in the following way: $\mathbf{momentum}_i = [\alpha \rho_L + (1-\alpha)\rho_G] \mathbf{u}_i $, where $\rho_L=1010$ and $\rho_G=1.18415$ are the liquid and gas densities, respectively. Liquid volume fraction is related to capture efficiency and is thus the most important quantity to predict for this dataset from a $\text{CO}_2$-capture perspective.

Node type and contact angle information are not explicitly provided in the data, but may be derived from boundary index information as indicated in Table \ref{tab:var}.

\begin{table}
\begin{tabular}{lrrrr}
\toprule
          Node Type & 2D Data Boundary Index & 3D Data Boundary Index & Contact Angle \\
\midrule
                 fluid &  1.79769313486232e+308 &  1.79769313486232e+308 &           N/A \\
          liquid\_inlet &                     34 &                     47 &           N/A \\
               outlet &                     31 &                     46 &           N/A \\
            gas\_inlet &                     30 &                     45 &           N/A \\
  inlet\_to\_outlet\_wall &                     29 &                     44 &           180 \\
            side\_wall &                     28 &                     43 &          33.5 \\
              packing &                     27 &                     42 &          33.5 \\
\bottomrule
\end{tabular}
\caption{The C3FD data's mapping from boundary index to node type and contact angle.}
\label{tab:var}
\end{table}

\section{Experiment details}
\label{sec:experiment_details}

\subsection{\CTFD experiment details}
For \CTFD, the surrogate inputs are: momentum, liquid volume fraction, contact angle, and node type. We also included the same edge features as \citet{pfaff2020learning}.The surrogate must predict changes in the momentum and liquid volume fraction, while the other variables are constant at each node. We observed better prediction performance when excluding pressure as a model input and output. 

For MGN input noise, we used $0.02$ for horizontal momentum ($x$-direction in 2D; $x$-, $y$-directions in 3D), $0.03$ for vertical momentum ($y$-direction in 2D; $z$-direction in 3D), and $0.05$ for liquid volume fraction. These values were not heavily optimized, and preliminary experiments suggest performance benefits from using larger values.

\citet{pfaff2020learning} found that 15 MP steps provided good performance. We also used 15 MP steps in our study, except in our 3D experiments (3 MP steps) and in our higher-order integrator experiments (indicated in the plot legends). As with \citet{pfaff2020learning}, we used 128 latent dimensions to encode nodes and edges for the 2D dataset. We used 150 latent dimensions for the 3D dataset.

As in \citet{pfaff2020learning}, we trained models with the Adam optimizer and an exponentially decaying learning rate, decaying from \num{1e-3} to \num{1e-7} over 4,000,000 steps. We typically stopped training at step 400,000 or earlier, when performance plateaued, as shown in our plots.

When using patch training, we randomly sampled patches from across different simulations and  timesteps. For 2D training, we created 12 patches (3 evenly-spaced partitions along the 100mm width, 4 along the 200mm height). For 3D training, we created 2,160 patches (12 evenly-spaced partitions along each horizontal axis, 15 along the vertical axis), 1,980 of which were non-empty. Each 2D and 3D patch contained on the order of \num{1e4} nodes, on average. Future work may achieve better 3D load balancing by using a more sophisticated decomposition scheme than uniform space partitioning.

On 3D experiments with patch training, we used 32 V100 16 GB GPUs, each with a batch size of 1 patch. On 2D experiments with patch training, we used 12 V100 16 GB GPUs, each with a batch size of 1 patch. Since 2D \CTFD's computational domain was small enough to fit on certain GPUs, we also used one A100 80 GB GPU with a batch size of 1 full domain to compare. We note that with 2D \CTFD, MGN weight updates used a roughly equal number of nodes for patch and non-patch training.

Validation errors were based on 1,000 randomly selected full-domain examples, except on 2D and 3D patch training experiments where we computed errors based on 1,000 patches.

\subsection{CylinderFlow experiment details}

We used 8 V100 16 GB GPUs to run experiments on the CylinderFlow dataset \citep{pfaff2020learning}. We used a batch size of 8 for each GPU (for an effective batch size of 64). For patch training, we also partitioned the grid into four equal-size subdomains. Following \citet{pfaff2020learning}, we use 15 MP steps, 128 latent dimensions to encode nodes and edges, and node type and momentum as inputs with $0.02$ input noise. We also used the same edge features. We predicted momentum and used the same learning rate schedule approach we described for \CTFD data, stopping training when loss plateaus. 

\section{Supporting Results}
\label{sec:supporting}

\subsection{Rollouts}

\begin{figure}[h]
\centering
\begin{subfigure}{.24\textwidth}
  \centering
  \includegraphics[width=\textwidth]{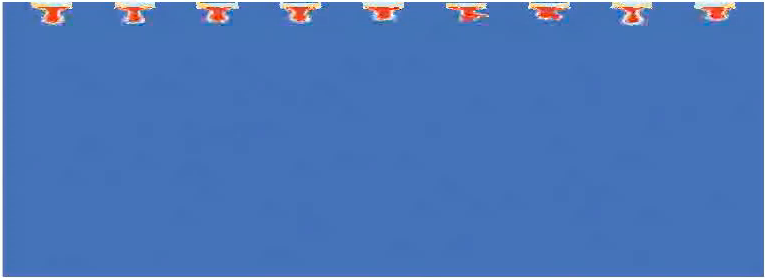}
\end{subfigure}%
\begin{subfigure}{.24\textwidth}
  \centering
  \includegraphics[width=\textwidth]{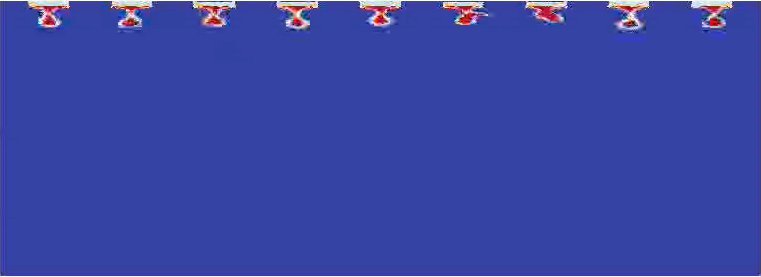}
\end{subfigure}%
\begin{subfigure}{.24\textwidth}
  \centering
  \includegraphics[width=\textwidth]{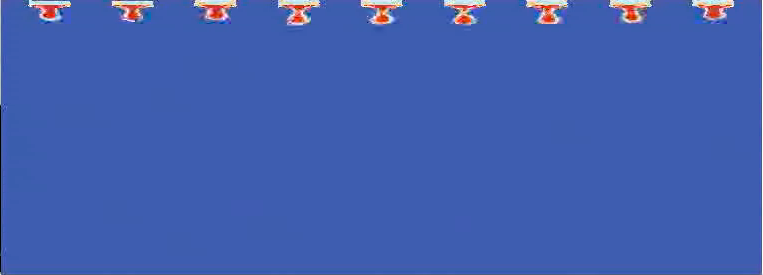}
\end{subfigure}%
\begin{subfigure}{.24\textwidth}
  \centering
  \includegraphics[width=\textwidth]{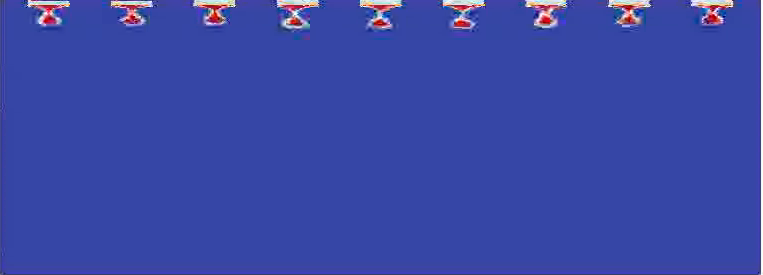}
\end{subfigure}%
\\
\begin{subfigure}{.24\textwidth}
  \centering
  \includegraphics[width=\textwidth]{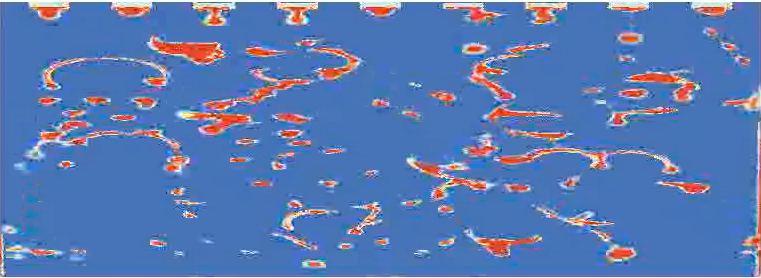}
\end{subfigure}%
\begin{subfigure}{.24\textwidth}
  \centering
  \includegraphics[width=\textwidth]{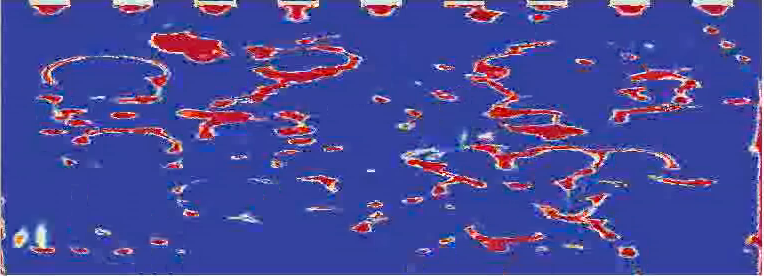}
\end{subfigure}%
\begin{subfigure}{.24\textwidth}
  \centering
  \includegraphics[width=\textwidth]{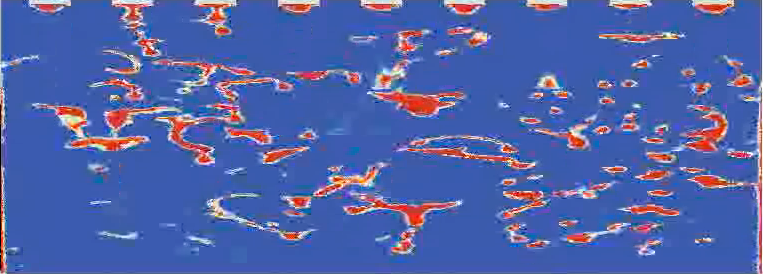}
\end{subfigure}%
\begin{subfigure}{.24\textwidth}
  \centering
  \includegraphics[width=\textwidth]{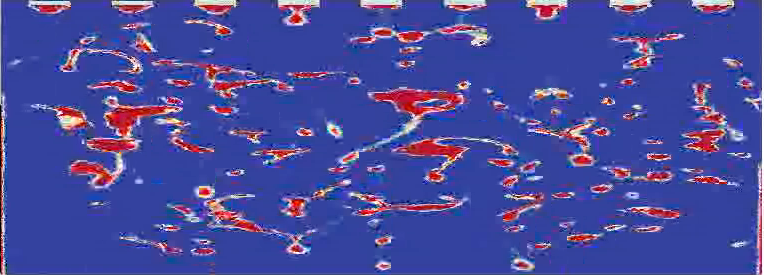}
\end{subfigure}%
\\
\begin{subfigure}{.24\textwidth}
  \centering
  \includegraphics[width=\textwidth]{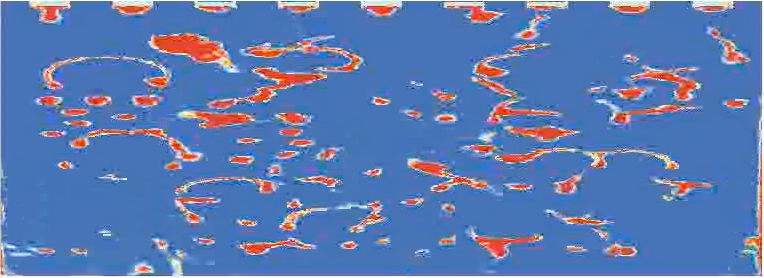}
\end{subfigure}%
\begin{subfigure}{.24\textwidth}
  \centering
  \includegraphics[width=\textwidth]{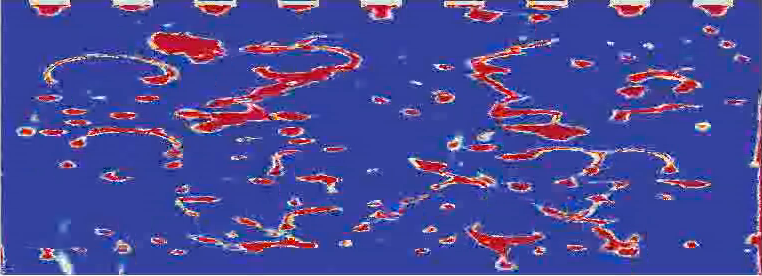}
\end{subfigure}%
\begin{subfigure}{.24\textwidth}
  \centering
  \includegraphics[width=\textwidth]{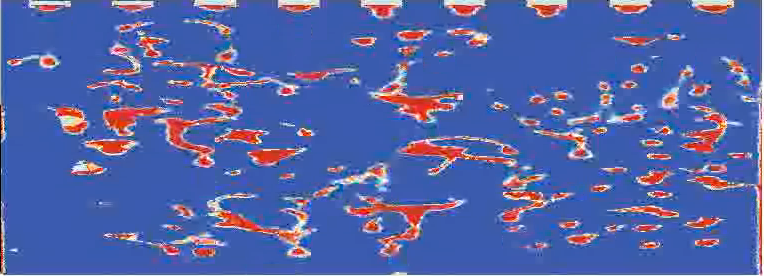}
\end{subfigure}%
\begin{subfigure}{.24\textwidth}
  \centering
  \includegraphics[width=\textwidth]{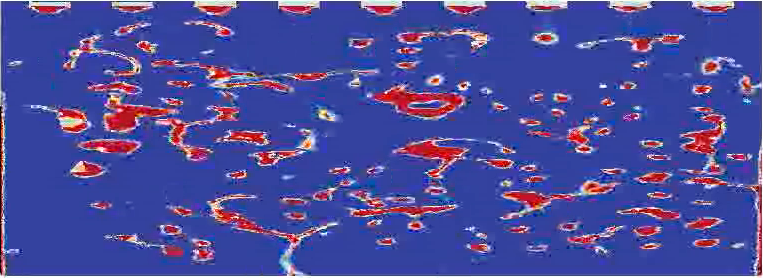}
\end{subfigure}%
\caption{\textbf{MGN has stable and visually accurate performance across long time horizons on domains with more than 100K nodes, even when generalizing to a new domain.} The top, middle, and bottom rows correspond to timesteps 30, 250, and 500. From left to right, the columns represent: the MGN's rollout on the 2D \CTFD domain, the ground truth CFD simulation for this domain, the MGN's rollout on a new 2D domain (the column's packings are configured differently), and the ground truth CFD simulation on this new domain. Liquid volume fraction is visualized.}
\label{fig:rollout}
\end{figure}

Figure~\ref{fig:rollout} shows several frames of the rollouts predicted with the MGN that used 15 MP steps and FE updates (discussed in Section~\ref{sec:int_exp} and Figure~\ref{fig:heun}).
After training on 2D \CTFD, we evaluated this model on a test simulation with an intermediate velocity (testing velocity interpolation). Additionally, we evaluated it on a new 2D domain created by taking a new cross-section of the 3D \CTFD domain.

\subsection{Training on a large domain is better than generalizing to it}
\label{sec:gen}

An alternative to training on a large domain via domain decomposition is simply training on a smaller domain. For instance, \citet{pfaff2020learning} showed that MGNs can generalize from training meshes with roughly \num{2e3} nodes to testing meshes with roughly \num{2e4} nodes. However, it is unclear if relying on such generalization is sufficient, or if we need to train on domains with scales and shapes similar to those of the test simulations. Preliminary results on this subject suggest that changing the shape of the domain can cause a small but notable amount of harm to performance \citep{pfaff2020learning}. 

\begin{figure}[h]
\centering
\begin{subfigure}{.48\textwidth}
  \centering
  \includegraphics[width=\textwidth]{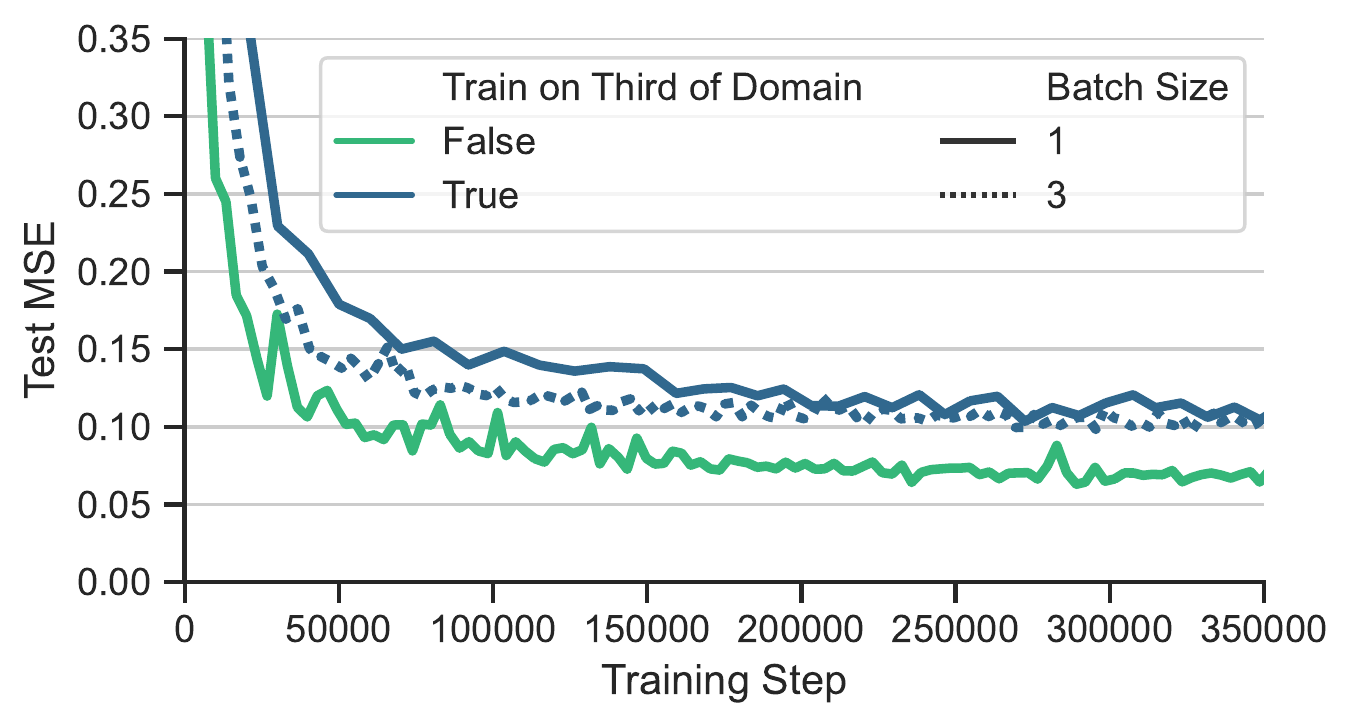}
\end{subfigure}
\hfill
\begin{subfigure}{.505\textwidth}
  \centering
  \includegraphics[width=\textwidth]{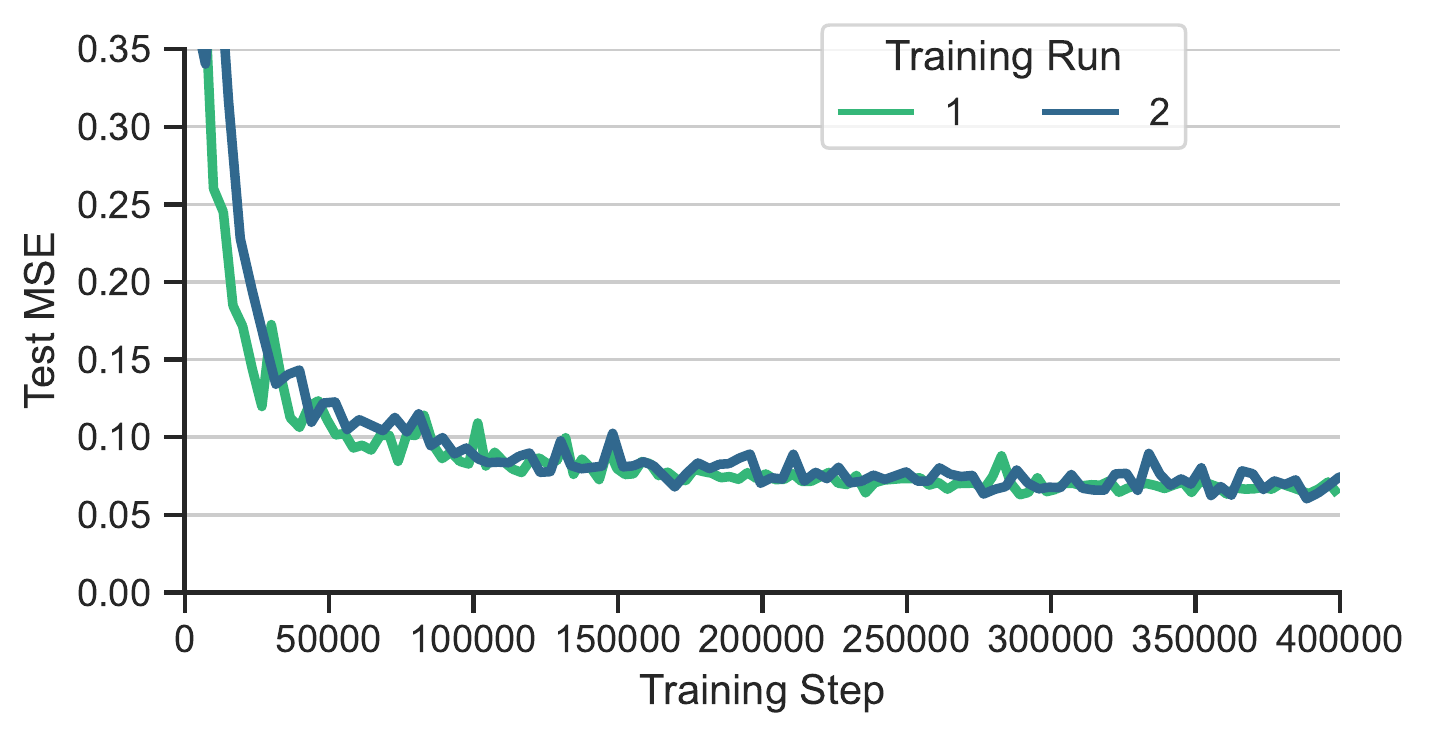}
\end{subfigure}
\caption{\textbf{Generalizing from a smaller mesh is inferior to training on the full domain}. Training only on the left third of our 150K node 2D \CTFD domain and then generalizing at test time leads to worse test performance (left), regardless of whether or not the batch size is adjusted to account for training on a subdomain. The baseline model's training has low variance across runs (right)---the standard deviation of the minimum test RMSEs reached by the two baseline models is 0.003. Relative to this standard deviation, the gap (left) between the minimum test RMSEs of generalizing from a smaller mesh and the baseline approach is large at 0.036.}
\label{fig:subset}
\end{figure}

To motivate the need for patch training, Figure~\ref{fig:subset} shows that training on the left third of the 2D \CTFD domain yields inferior test performance than training on the full domain. Patch training can address this by making it possible to train on the entire domain, even when it does not fit on an accelerator.

\subsection{Patch training is an effective way to train on a large domain}

\begin{figure}[h]
\centering
\begin{subfigure}{.48\textwidth}
  \centering
  \includegraphics[width=\textwidth]{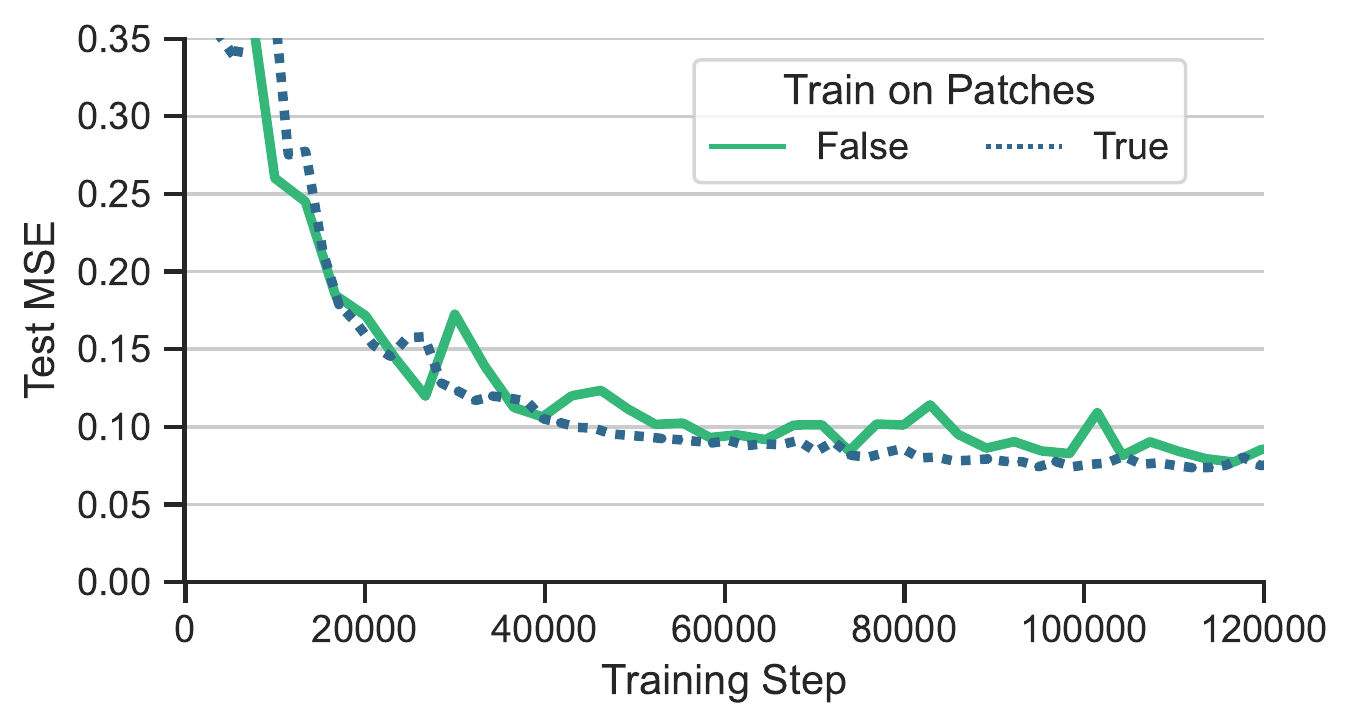}
\end{subfigure}
\hfill
\begin{subfigure}{.49\textwidth}
  \centering
  \includegraphics[width=\textwidth]{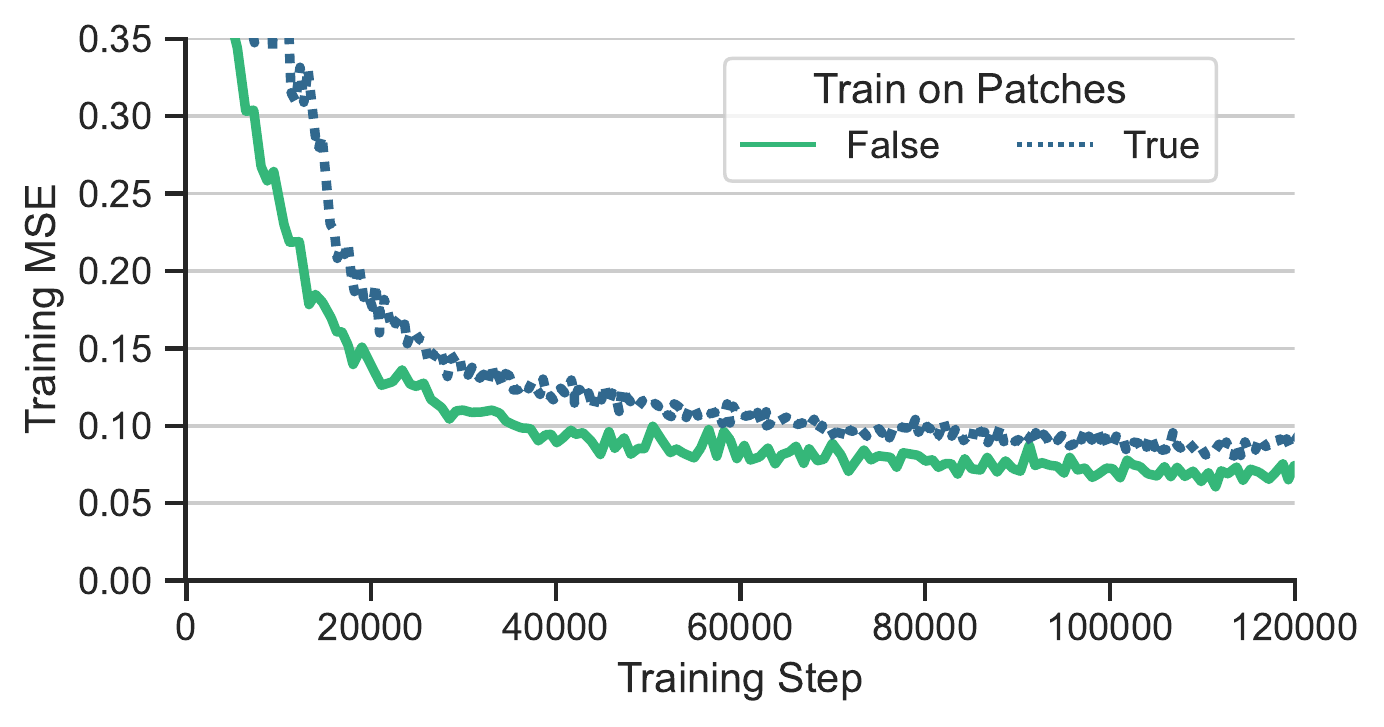}
\end{subfigure}
\caption{\textbf{Patch training can replace regular training on the full domain}. Training on patches produced by domain decomposition leads to similar test (left) and training (right) performance, even though patch training uses randomly sampled patches from different simulations and timesteps. This illustrates that domain decomposition can facilitate training on domains that would normally cause out-of-memory errors. Experiments run on 2D \CTFD simulations with 150K nodes per timestep.}
\label{fig:patch}
\end{figure}

\begin{figure}[h]
\centering
\begin{subfigure}{.46\textwidth}
  \centering
  \includegraphics[width=\textwidth]{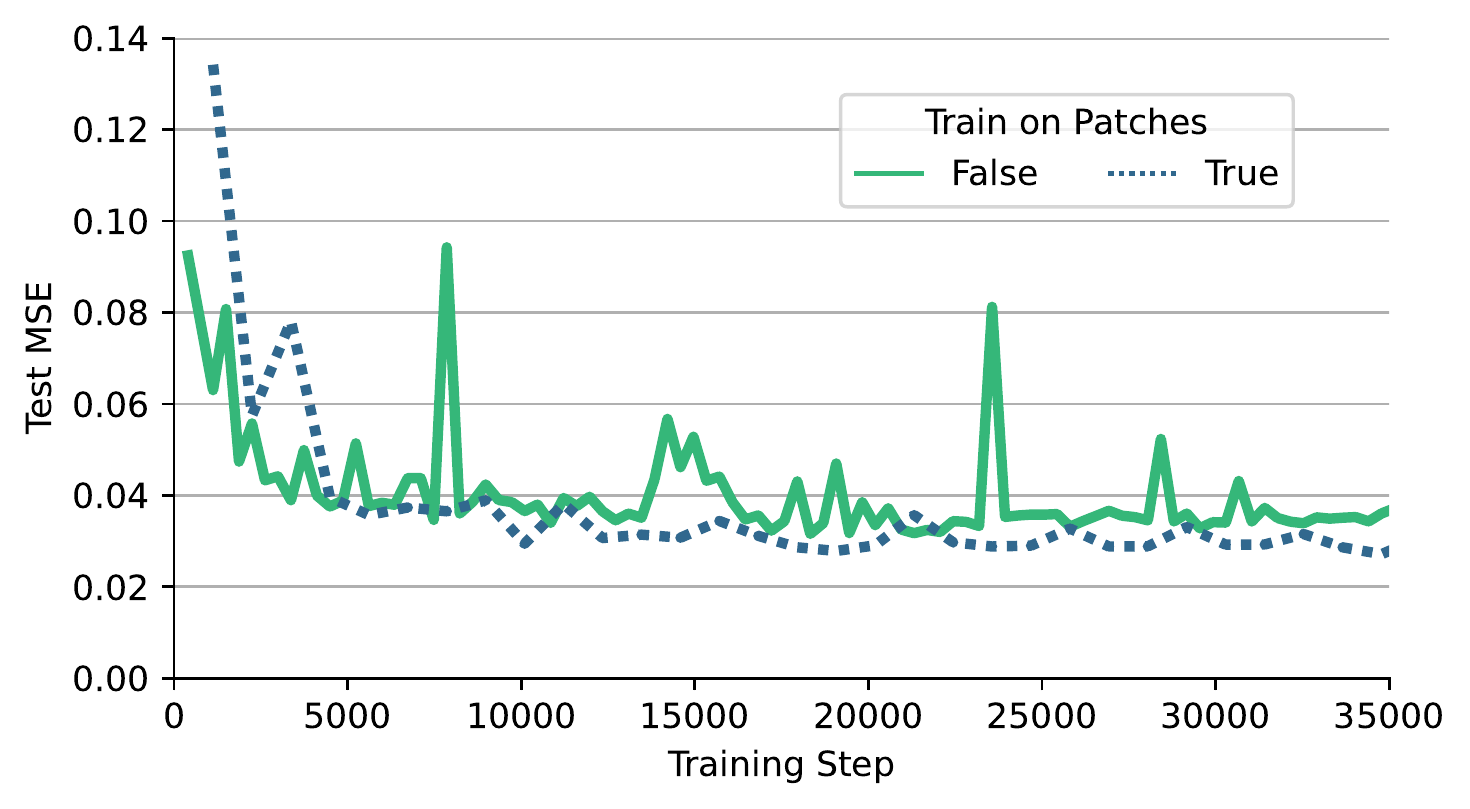}
\end{subfigure}
\hfill
\begin{subfigure}{.46\textwidth}
  \centering
  \includegraphics[width=\textwidth]{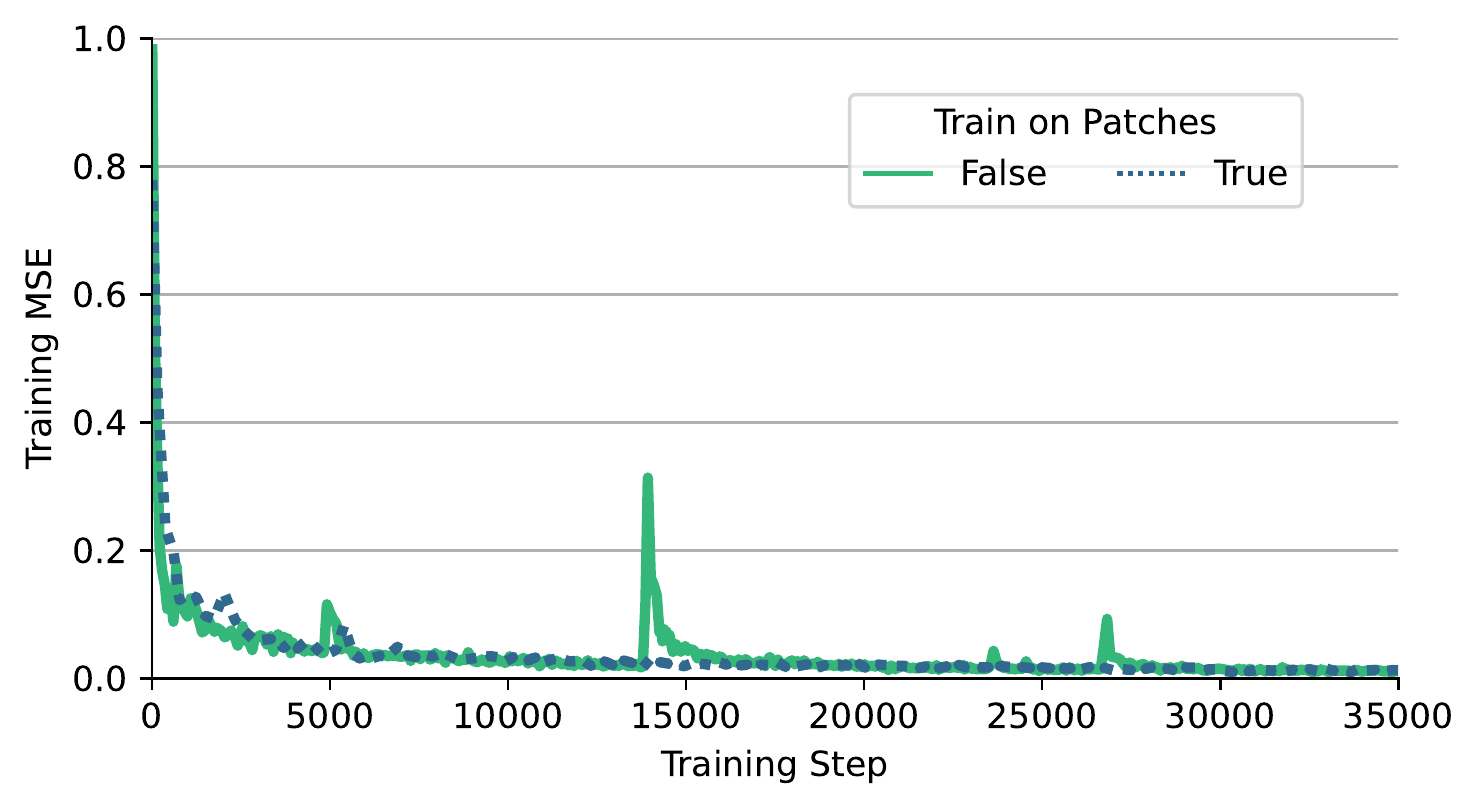}
\end{subfigure}
\caption{\textbf{Patch training is effective on CylinderFlow}. Patch training on the CylinderFlow simulations improves the loss on the test (left) and training (right) datasets.}
\label{fig:cf}
\end{figure}

We show that patch training, which updates weights using a random sample of patches from different simulations and timesteps, produces error that is comparable to regular training by updating weights using the entire domain. This is true for both \CTFD (Figure~\ref{fig:patch}) and CylinderFlow (Figure~\ref{fig:cf}).

\subsection{Additional results on higher-order integration}
\label{sec:heun_train_and_third}

\begin{figure}[h]
\centering
\begin{subfigure}{.32\textwidth}
  \centering
  \includegraphics[width=\textwidth]{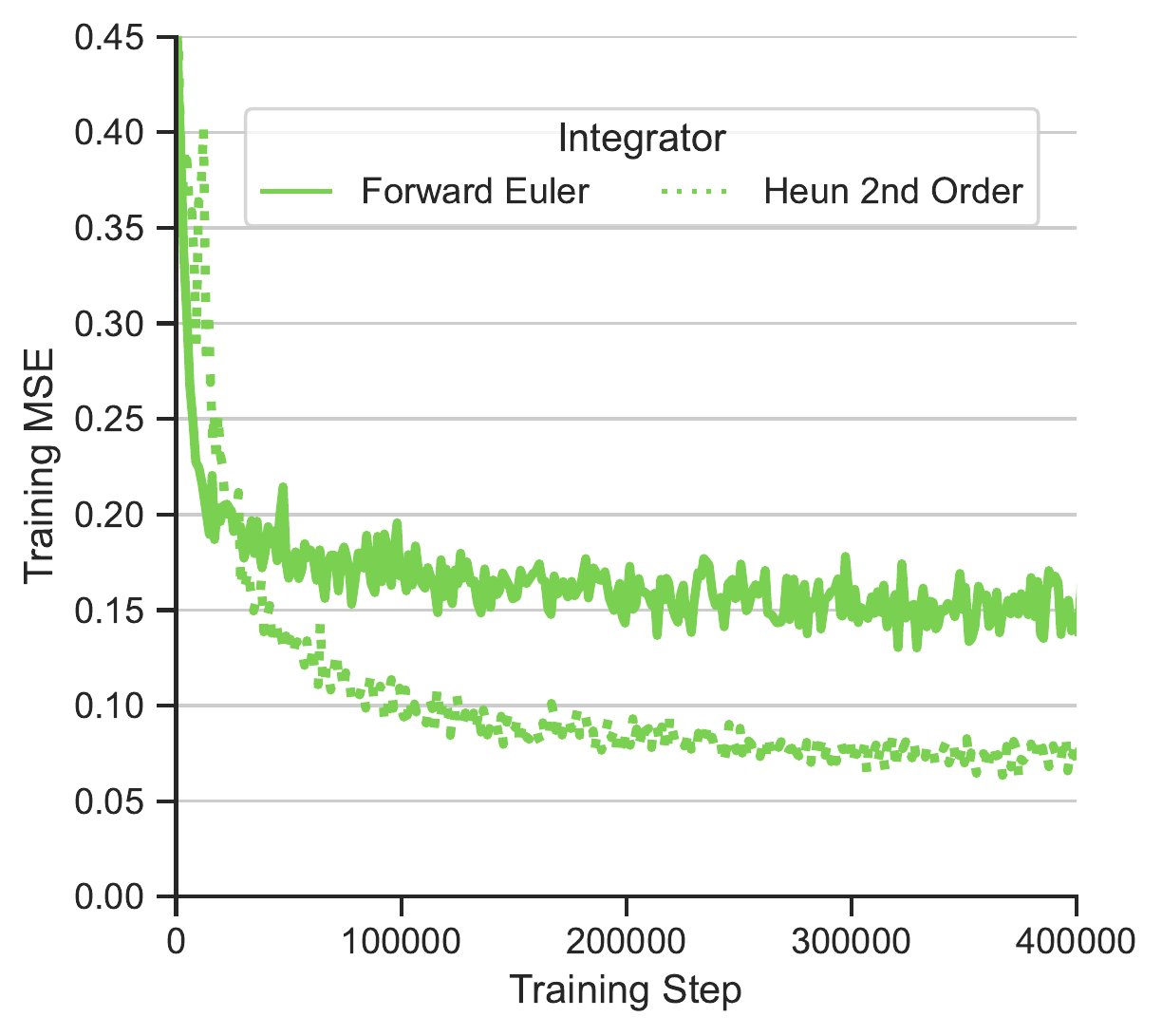}
\end{subfigure}
\hfill
\begin{subfigure}{.323\textwidth}
  \centering
  \includegraphics[width=\textwidth]{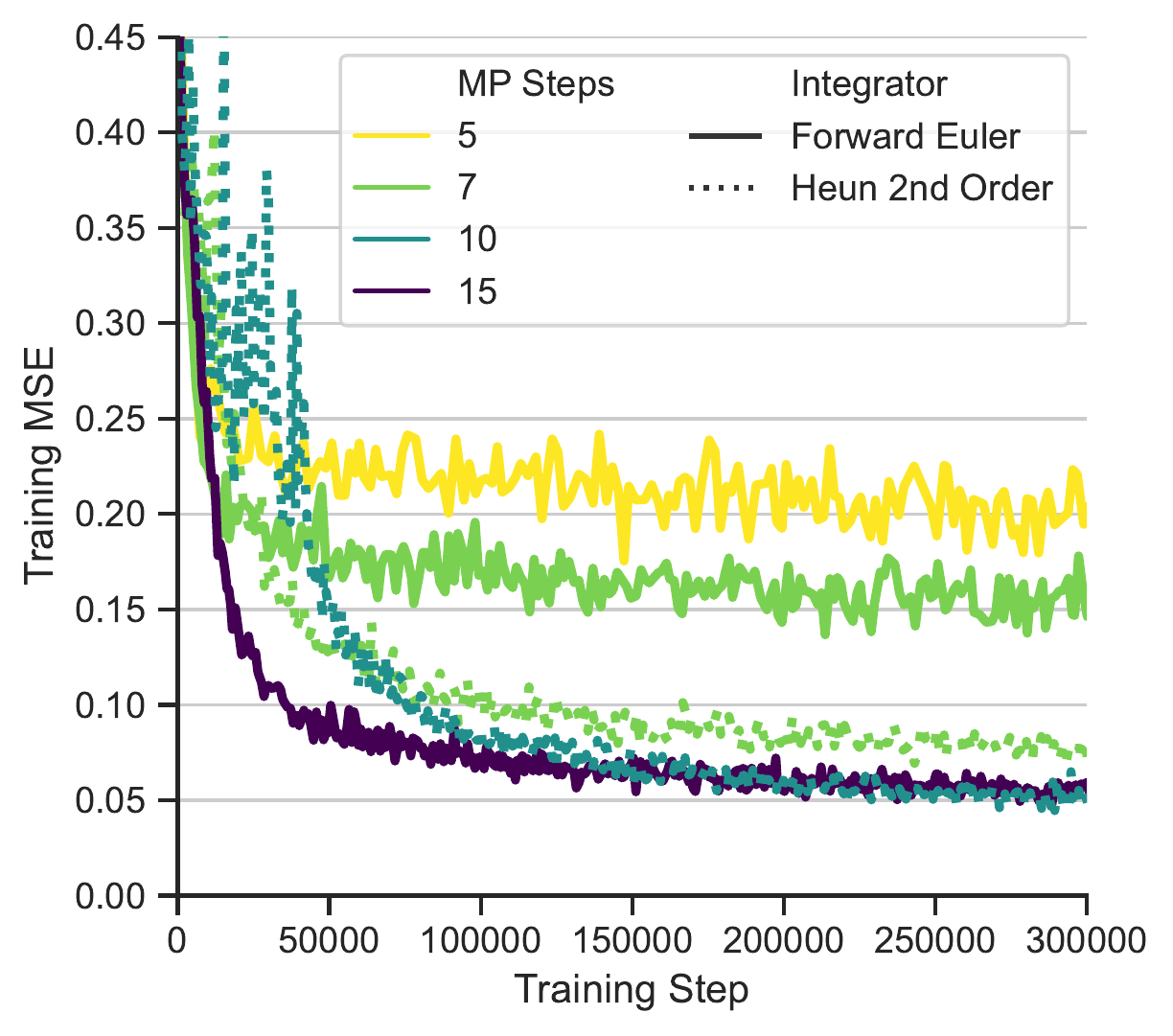}
\end{subfigure}
\hfill
\begin{subfigure}{.314\textwidth}
  \centering
  \includegraphics[width=\textwidth]{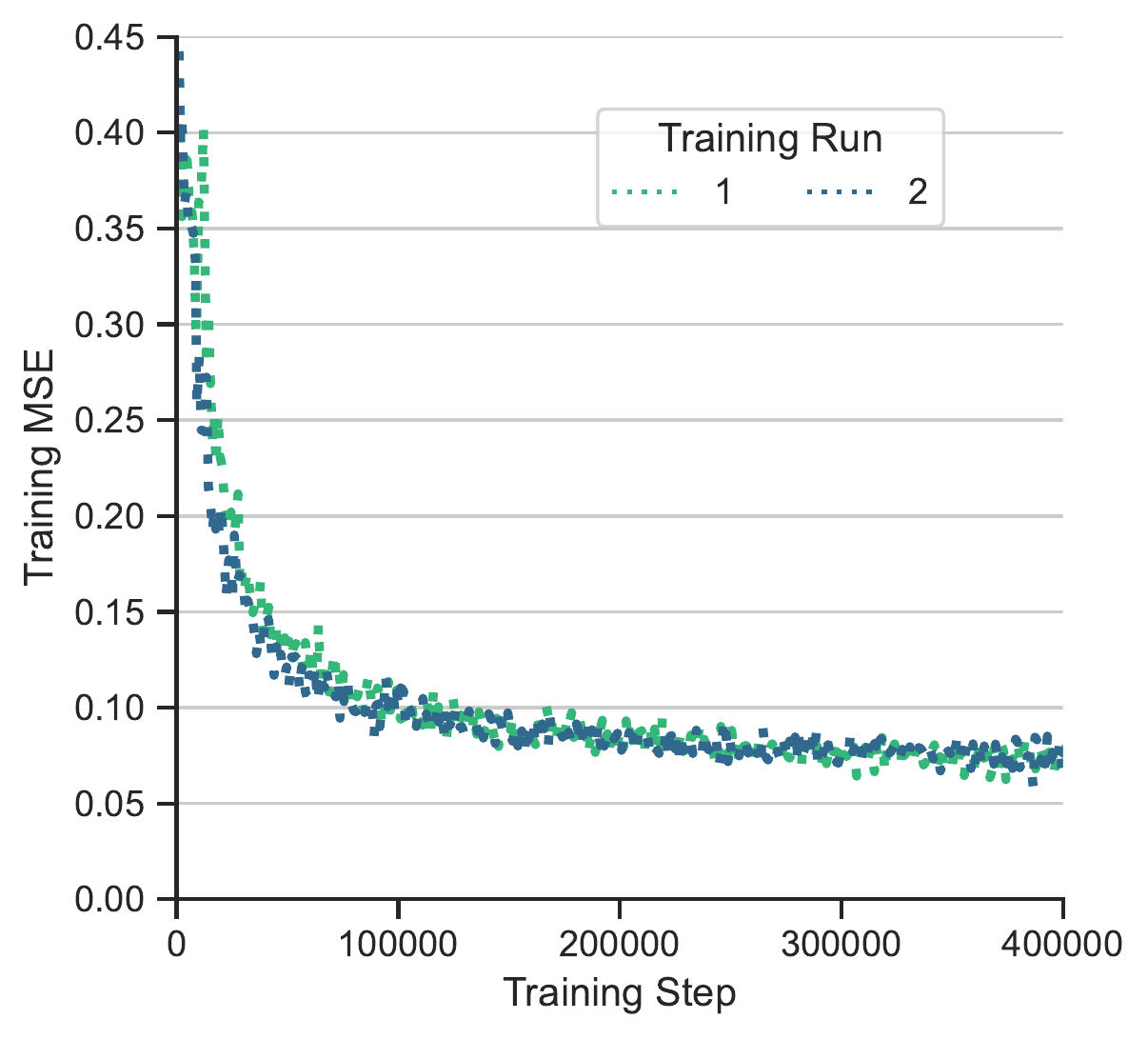}
\end{subfigure}
\caption{\textbf{Higher-order integration enhances learning on training data}. (\textit{Left}) Using Heun's second-order (H2) method on an MGN with 7 MP steps halves the training error, relative to the baseline forward Euler (FE) method. (\textit{Middle}) H2 and 7 MP steps is competitive with FE and 15 MP steps, despite the visible (and previously documented) benefits of more MP steps. (\textit{Right}) The benefits of higher-order integration are stable across two training runs. Experiments run on 2D \CTFD.}
\label{fig:heun_train}
\end{figure}

\begin{figure}[t]
\centering
\begin{subfigure}{.5\textwidth}
  \centering
  \includegraphics[width=\textwidth]{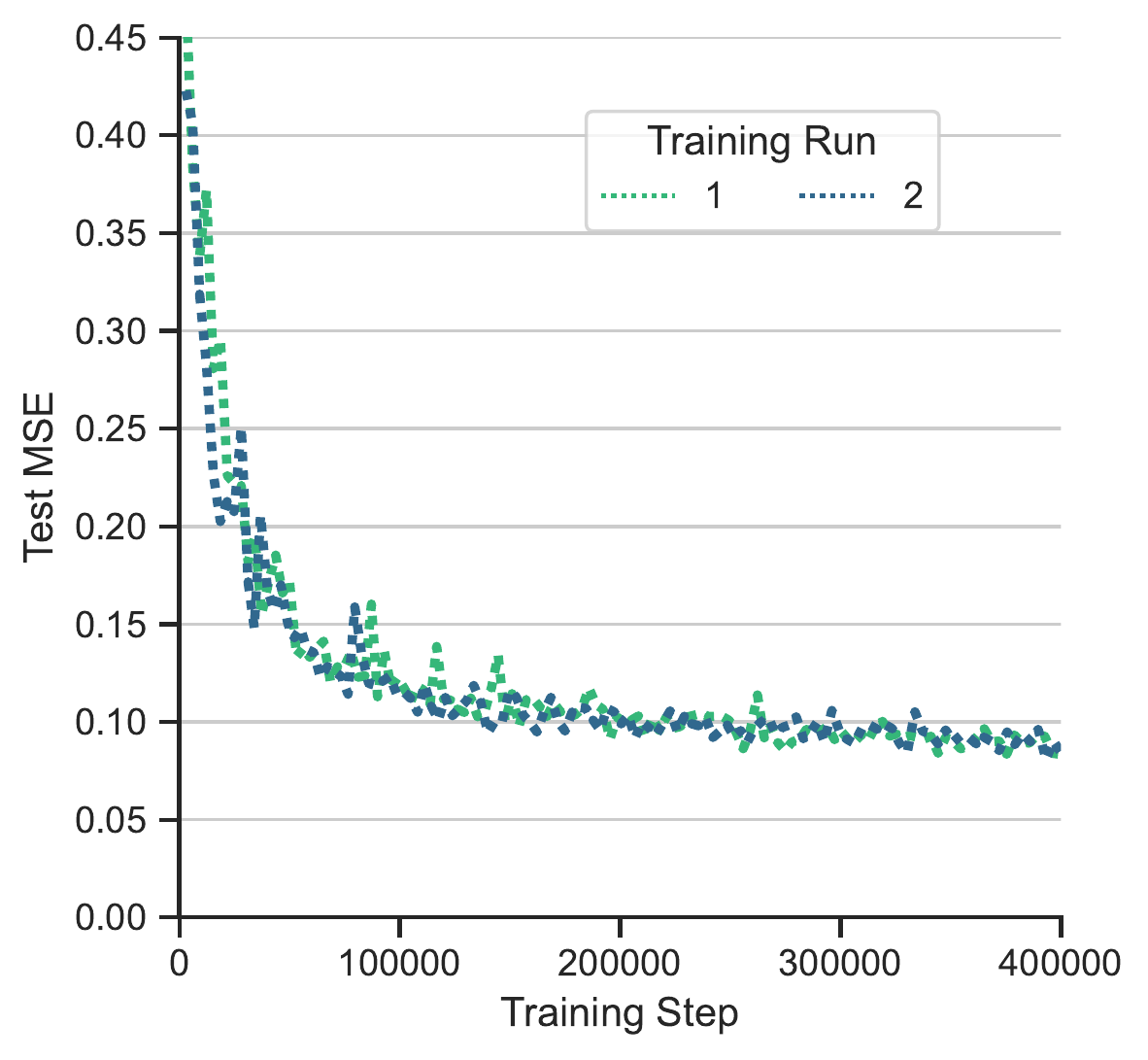}
\end{subfigure}%
\caption{\textbf{Higher-order integration's benefit is stable across training runs.}}
\label{fig:int_stable}
\end{figure}

\begin{figure}[h]
\centering
\begin{subfigure}{.493\textwidth}
  \centering
  \includegraphics[width=\textwidth]{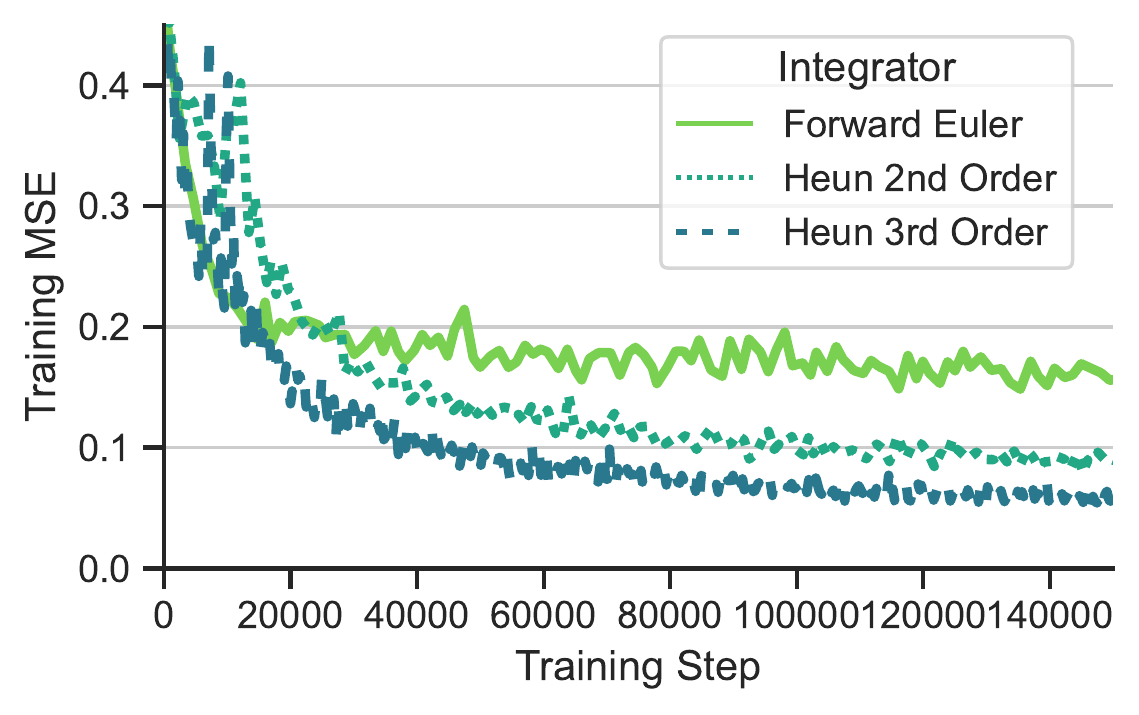}
\end{subfigure}
\hfill
\begin{subfigure}{.48\textwidth}
  \centering
  \includegraphics[width=\textwidth]{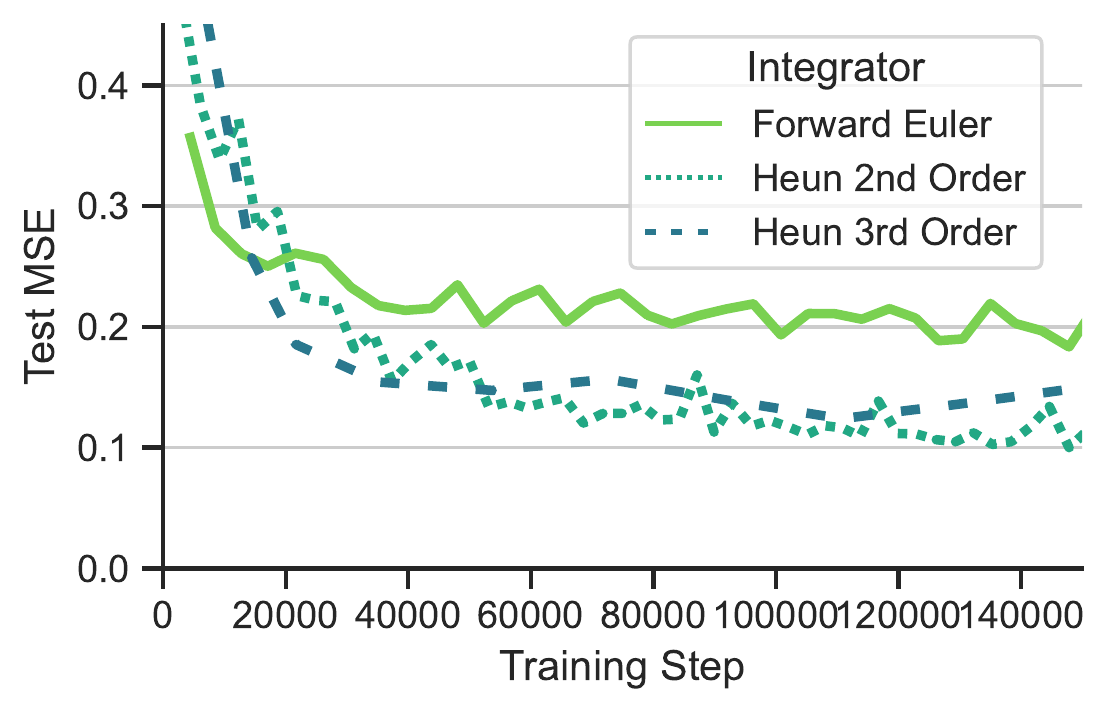}
\end{subfigure}
\caption{\textbf{Third-order integration improves training but not test performance}. All MGN models use 7 MP steps.}
\label{fig:heun_third}
\end{figure}

In Figure~\ref{fig:heun}, we observed that Heun's second-order method benefits test errors; Figure~\ref{fig:int_stable} shows that this benefit is stable across training runs. In Figure~\ref{fig:heun_train}, we also show that this benefit also extends to training. Notably, with Heun's third-order method, training error decreases even faster (see Figure~\ref{fig:heun_third}). This is consistent with the hypothesis that mesh-based GNNs like MGN have difficulty removing not just $\mathcal{O}(h^2)$ but also $\mathcal{O}(h^3)$ local truncation errors. However, we also observe test error is not best served by a third-order integrator, which suggests that higher order integrators would benefit from additional regularization. In other words, irremovable/unlearnable local truncation errors may provide regularization during training, and their removal by higher-order integrators may create a need for more regularization for good test performance.

\section{Related work}
\label{sec:related}
Neural network surrogates for similar large-scale CFD simulations have been studied previously \citep{Bartoldson2022}, but we are unaware of prior use of physically faithful GNN surrogates that generalize across domains in such contexts. In particular, \citet{pfaff2020learning} demonstrated that MeshGraphNets (MGN) can learn generalizable rules for updating physics fields from meshes, and MultiScale MGNs attained state-of-the-art performance on many tasks \citep{fortunato2022multiscale}, but their memory inefficiencies prevented their use on large, application-relevant graphs. 

More broadly, scaling GNNs to large graphs is an important topic that has been approached in various ways, including parallelism and off-accelerator memory use \citep{Jia2020} and more efficient learning algorithms. Prior algorithmic work has improved operating efficiency on large graphs by sampling \citep{hamilton2017inductive,chiang2019cluster} or 
otherwise modifying the function computed 
by an inefficient GNN \citep{zhou2021brief}---such approaches may be critical when average node degree is high. In contrast, we focused on domains that demand physical faithfulness, making approximations less attractive.\footnote{If the state of node $u$ affects the state of node $v$ according to physical laws, then a surrogate should 
seemingly not learn 
an update rule for $v$ based on a sampling algorithm that may exclude $u$. Indeed, while compatible with other efficiency methods, we do not integrate sampling-based approaches into our study to avoid harming the inductive biases of the surrogates we consider.} Accordingly, we take advantage of the manageable node degree associated with physics meshes to introduce a GNN physics surrogate scaling approach that can provide training mathematically equivalent to training on a mesh that would normally cause out-of-memory errors.

Prior to our work, \citet{EAGLE} introduced EAGLE, a benchmark dataset for physics surrogate learning with over 1,000 2D simulations. EAGLE contains meshes with 3,388 nodes per timestep on average, while our 3D \CTFD dataset uses a mesh with 3.1M nodes at each timestep ($\sim 1000$x larger). Further differences include the following: our datasets use the same mesh as the CFD solver, while EAGLE downsamples the CFD mesh greatly; we consider a static mesh rather than a dynamic mesh; our CFD simulations involve $\text{CO}_2$-capture columns relevant to understanding capture performance as a function of simulation parameters; and our geometries include a variety of complex obstacles in the form of column packings. Separately, \citet{EAGLE} showed that the number of message-passing iterations can be reduced by using attention and graph pooling. In contrast, we showed that the number of message-passing iterations can be reduced by using a higher-order integrator. 

Mesh-based graph surrogates have previously been used to produce time derivative estimates for use in a higher-order integration scheme \citep{lienen2022learning}. However, this prior work used GNNs that were embedded within a differential equation solving framework (a finite element method), which grants structure to the rules learned by the GNN that is not available when using the more general mesh surrogate GNNs that are our focus. Further, this prior work used higher-order integration as a byproduct of inserting a GNN into a differential equation solver, which runs for several timesteps. In contrast, we motivated consideration of higher-order integration through analysis of its effect on the learning task and inductive biases of any network that predicts time derivatives, and we tested related hypotheses by comparing higher-order integration to a baseline first-order method.

\section{Limitations and Future Work}
\label{sec:limitations}

Our work only scratches the surface of opportunities for SCALES2 to improve machine learning of GNN surrogates that operate on meshes. For example, it may be possible to obtain a higher-order integrator's benefits without incurring its memory and compute overhead---the Predict–Evaluate–Correct (PEC) framework may facilitate this at inference time.\footnote{We note that the memory and compute overhead of higher-order integration may be addressed by lowering the MP steps: for example, FE with 15 MP steps has roughly the same memory cost and performance as Heun's second-order method with 7 MP steps.} Also, better performance may be seen with other integrators, such as Ralston's methods, to minimize the local truncation error \citep{ralston1962runge}. Relatedly, our results indicate that training details (such as regularization strength) may need to be adjusted to match the integrator's order.

Importantly, given a number of MP steps, a higher-order integrator leads to more computation on the accelerator and (if computation on the accelerator is a bottleneck) longer training steps. However, it can also reduce the number of training steps needed to reach a target accuracy. Indeed, while we do not thoroughly characterize speedups associated with higher-order integration, there were parameter settings where training speedups resulted from its use. In particular, our 3D models trained with H2 reached target error levels in strikingly fewer steps than FE models (see Figure~\ref{fig:3d}) while having the same time per training step ($\approx 1.05$ seconds): e.g., models beat 0.33 test MSE in about 12 hours ($\approx 40K$ steps) with H2 compared to about 54 hours ($\approx 185K$ steps) with FE. 

Another limitation of our work is that we used 32 accelerators to train on the 3D \CTFD dataset quickly with a large effective batch size (32 patches). Future work may consider using smaller batch sizes (perhaps with correspondingly smaller learning rates) to reduce the relevance of accelerator count to training MGN-like models on very large computational domains such as those in \CTFD.

\begin{figure}[t]
\centering
\begin{subfigure}{.7\textwidth}
  \centering
  \includegraphics[width=\textwidth]{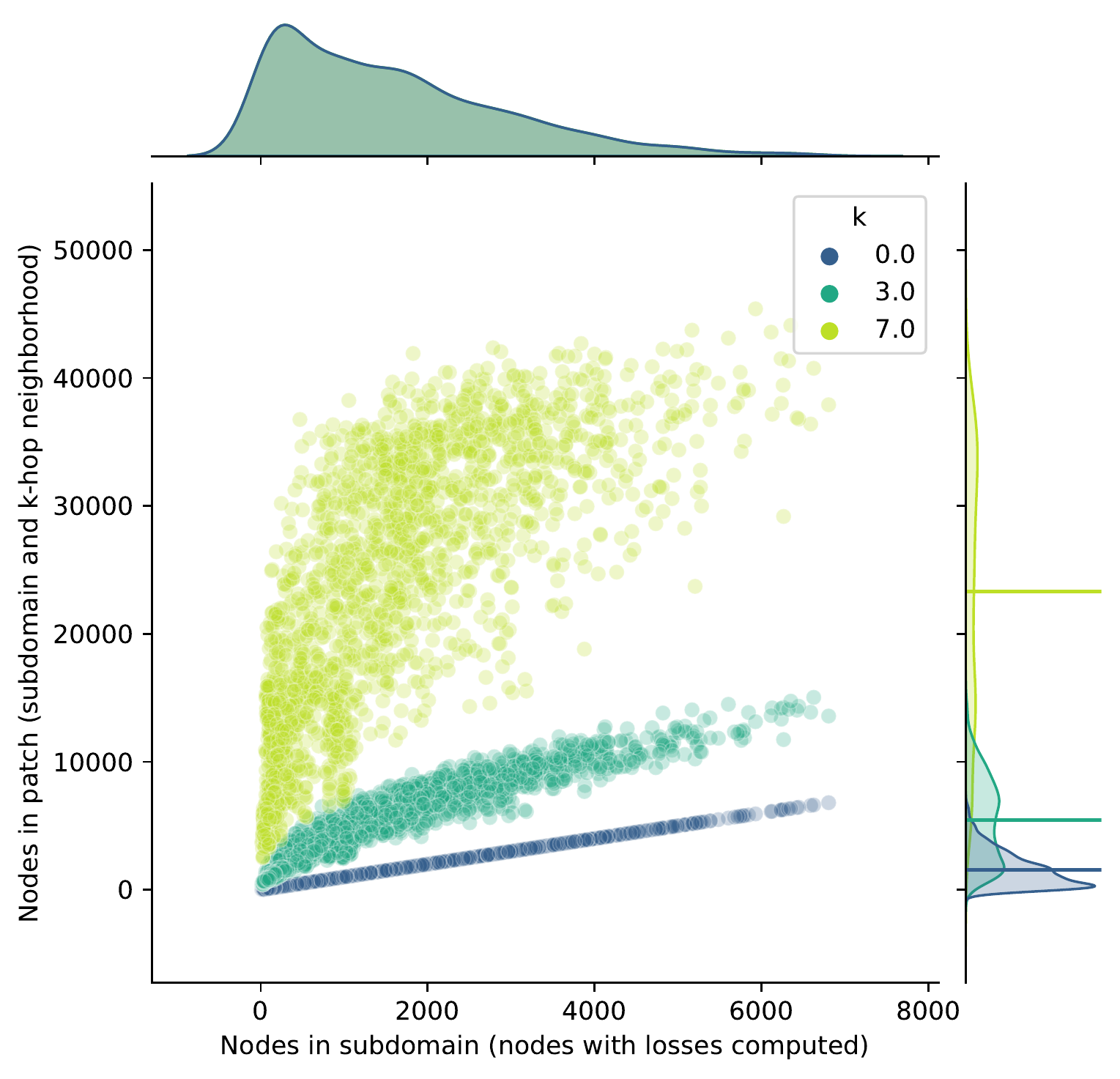}
\end{subfigure}%
\caption{\textbf{With 3D C3FD, increasing $k$ leads to increasingly larger patches.} Each patch contains its subdomain's $k$-hop neighborhood. For each $k$, horizontal bars on the right denote the average number of nodes in a patch, which increases nonlinearly with $k$. Thus, as the number of MP steps $m$ increases, satisfying condition 1 from Section \ref{sec:DD} ($m \leq k$) requires increasingly more resources.}
\label{fig:patch_scaling}
\end{figure}

Relatedly, more sophisticated domain decomposition approaches are likely to provide benefits. Our approach constructs \patches, each of which contains a subdomain with a \ghostZone{} border to facilitate training that is mathematically equivalent to training on the full domain directly. Training on a \patch{} is more compute-intensive than training on just the subdomain. Moreover, on our 3D meshes, average node degree is higher than on 2D meshes (using edges created via Delaunay triangulation). This leads to faster growth in subdomains' $k$-hop neighborhoods and, using 2,160 3D patches, out-of-memory errors on a 16GB GPU when going beyond \thickness{k} $=7$ (see Figure \ref{fig:patch_scaling} for a depiction of this growth). Since larger $k$-hop neighborhoods may be desired without resorting to use of more accelerator memory or creation of more but smaller patches, future work may consider creating edges with a nearest neighbors algorithm to tame node degree. Alternatively, a partitioning approach that balances the node counts of the resulting \patches{} rather than subdomain spatial extents may help. Finally, this issue may also be addressed by violating condition 1 (i.e., by setting $k$ to be less than the MP steps $m$): preliminary experiments we ran on CylinderFlow suggest that performance is robust to using smaller $k$, provided that a majority of nodes in the subdomain have access to their true $m$-hop neighbors. We expect that further work on these issues will lead to better scaling and performing GNN surrogate simulators for mesh physics.

\end{document}